\documentclass[pdflatex,sn-mathphys-num]{sn-jnl}

\usepackage{graphicx}%
\usepackage{multirow}%
\usepackage{amsmath,amssymb,amsfonts}%
\usepackage{amsthm}%
\usepackage{mathrsfs}%
\usepackage[title]{appendix}%
\usepackage{xcolor}%
\usepackage{textcomp}%
\usepackage{manyfoot}%
\usepackage{booktabs}%
\usepackage{algorithm}%
\usepackage{algorithmicx}%
\usepackage{algpseudocode}%
\usepackage{listings}%
\usepackage{longtable}%
\usepackage{array}%
\usepackage{subcaption}
\usepackage{float}
\usepackage{caption}
\theoremstyle{thmstyleone}%
\theoremstyle{thmstyletwo}%

\theoremstyle{thmstylethree}%

\begin{document}

\title[Article Title]{Cross-Validated Cross-Channel Self-Attention and Denoising for Automatic Modulation Classification}


\author[1]{\fnm{Prakash} \sur{Suman}}\email{prakash.suman@gmail.com}

\author*[1]{\fnm{Yanzhen} \sur{Qu}}\email{yqu@coloradotech.edu}

\affil[1]{\orgdiv{Computer Science, Engineering, and Technology},
\orgname{Colorado Technical University},
\orgaddress{\street{1575 Garden of the Gods Road},
\city{Colorado Springs},
\postcode{80907},
\state{CO},
\country{USA}}}


\abstract{This study addresses a key limitation in deep learning Automatic Modulation Classification (AMC) models, which perform well at high signal-to-noise ratios (SNRs) but degrade under noisy conditions due to conventional feature extraction suppressing both discriminative structure and interference. The goal was to develop a feature-preserving denoising method that mitigates the loss of modulation class separation. A deep learning AMC model was proposed, incorporating a cross-channel self-attention block to capture dependencies between in-phase and quadrature components, along with dual-path deep residual shrinkage denoising blocks to suppress noise. Experiments using the RML2018.01a dataset employed stratified sampling across 24 modulation types and 26 SNR levels. Results showed that denoising depth strongly influences robustness at low and moderate SNRs. Compared to benchmark models PET-CGDNN, MCLDNN, and DAE, the proposed model achieved notable accuracy improvements across -8 dB to +2 dB SNR, with increases of 3\%, 2.3\%, and 14\%, respectively. Cross-validation confirmed the model’s robustness, yielding a mean accuracy of 62.6\%, macro-precision of 65.8\%, macro-recall of 62.6\%, and macro-F1 score of 62.9\%. The architecture advances interference-aware AMC by formalizing baseband modeling as orthogonal subproblems and introducing cross-channel attention as a generalized complex interaction operator, with ablations confirming the critical role of feature-preserving denoising for robustness at low-to-medium SNR.}




\maketitle

\section{Introduction}\label{sec1}

Robust automatic modulation classification (AMC) at low and moderate signal-to-noise ratio (SNR) improves the detection and characterization of incumbent and opportunistic users, thereby enabling tighter spectral occupancy in wireless communications. Robust AMC at low SNR can shorten the sensing window or operate with lower sensing power while increasing available data transmission and improving channel utilization. More accurate modulation recognition in congested environments enables inference-aware transmission scheduling, thereby improving spectral efficiency.
Enhanced SNR robustness reduces uncertainty in admission and handover decisions for cellular networks under congestion, especially when multiple spectrum bands exhibit heterogeneous interference \cite{tayyab_survey_2019}. Improved visibility into neighboring waveforms informs carrier aggregation choices that maximize throughput while minimizing mutual interference, improving user experience, and data throughput rates \cite{adamu_analysis_2024}. In shared unlicensed spectrum use, identifying co-channel users improved policy compliance and protected priority access \cite{parvini_spectrum_2023}. With higher fidelity at low SNR, spectrum sensing can operate with larger spatial reuse factors, enabling more aggressive spectrum resuse without unacceptable interference. Robust AMC models can improve spectrum efficiency, enabling more aggressive spectrum reuse without sacrificing reliability.This paper proposes a deep learning AMC model that is robust to low-to-moderate SNR levels.\newline
Unlike prior AMC approaches that implicitly learn I/Q interactions through convolutional filters or complex-valued representations, this study explicitly models I/Q coupling as a structured two-token interaction problem. The novelty lies not in individual components such as attention or shrinkage networks, but in their structured integration and interpretation as a principled representation-learning framework for complex signals.

\begin{itemize}
\item We formalize In-phase (I) and Quadrature (Q) phase components interaction as a two-token cross-channel self-attention operator, providing a structured alternative to convolutioal and complex-valued mixing.\newline
\item We use a self-learning, self-adaptive dual-path deep residual shrinkage network (DP-DRSN) with Garrote Thresholding \cite{gao_wavelet_1998} to denoise radio signals for AMC.\newline
\item Cross-validation of the proposed model yields a mean classification accuracy of 62.6\%, a macro precision of 65.8\%, a mean macro recall of 62.6\%, and a mean macro F1 score of 62.9\%.\newline
\item Relative to benchmark models, PET-CGDNN \cite{zhang_efficient_2021}, MCLDNN \cite{chang_multitask-learning-based_2022}, and DAE \cite{ke_real-time_2022}, the proposed AMC model achieved modulation classification accuracy gains across the -8 dB to +2dB SNR range, with improvements of 3\%, 2.3\%, and 14\% respectively.\newline
\item We derive transferable design principle from controlled ablations, identifying denoising depth as the primary determinant of low-SNR robustness.\newline
\end{itemize}

We trained and evaluated on the standard RML2018.01a dataset \cite{oshea_over--air_2018}, which has been widely used in prior AMC research.
This paper is organized as follows: Section 2 discusses related work, Section 3 presents the problem statement, hypothesis, and research question, Section 4 outlines the method of proposed AMC model architecture, Section 5 details results of experiment and analysis, Section 6 provides discussion of ablation study, limitations, and recommendations for future research, and Section 7 concludes the paper.

\section{Related Work}\label{sec1}
The literature review highlights distinct performance patterns across deep learning methodologies applied to AMC.\newline 
CNN-based models,\cite{shaik_automatic_2021}, \cite{shen_multi-subsampling_2023}, \cite{wang_multitask_2024}, \cite{li_lightweight_2023}, \cite{shi_improved_2022}, \cite{harper_learnable_2024}, \cite{huynh-the_mcnet_2020}, \cite{xue_micnet_nodate}, \cite{nisar_lightweight_2023}, \cite{harper_automatic_2023}, \cite{xiao_multiscale_2023}, \cite{singh_automatic_2025}, \cite{duan_multi-modal_2023}, \cite{wei_adaptive_2023}, \cite{cao_modulation_2023}, \cite{tekbiyik_robust_2020}, \cite{an_robust_2023}, \cite{guo_robust_2023}, \cite{le_performance_2022}, \cite{triaridis_mm-net_2024}, \cite{wu_automatic_2020}, \cite{wang_automatic_2023}, \cite{xiao_complex-valued_2023}  remain the most widely applied, with average classification accuracies ranging from the mid 50\% level on complex scenarios to approximately 65\% on RML2016.10a and RML2018.01a for lightweight designs such as LightMFFS \cite{li_lightweight_2023}. Larger CNNs with enhancements, including squeeze-and-excitation (SE) blocks and constellation features, generally improve robustness but still exhibit variability across datasets, underscoring the challenges of generalization.\newline 
RNN-based \cite{ke_real-time_2022}, \cite{ansari_novel_nodate} approaches, particularly LSTMs, demonstrate a strong ability in sequence modeling. Ke and Vikalo \cite{ke_real-time_2022}, a DAE-LSTM-based model, achieved a maximum accuracy of approximately 98\% on RML2018.01a despite modest average performance. This indicates high potential when temporal dependencies are critical. Transformer models \cite{li_complex-valued_2024}, \cite{su_robust_2022}, \cite{zheng_tmrn-glu_2022} stand out for efficiency and scalability. Studies such as SigFormer \cite{su_robust_2022} and CV-TRN \cite{li_complex-valued_2024} achieved approximately 65\% accuracy across RML2016.10a, RML2016.10b, and RML2018.01a with parameter counts as low as 44k, underscoring their capacity to capture long-range dependencies with fewer resources than CNNs.\newline 
SNN models \cite{lin_fast_2024}, \cite{guo_end--end_2024} remain exploratory with relatively low accuracies (36–64\%) across RML2016.10a, RML2016.10b, and RML2018.01a, but their potential for energy efficiency warrants further research.\newline
Hybrid models \cite{zhang_efficient_2021}, \cite{suman_lightweight_2025}, \cite{tianshu_iq_2022}, \cite{gao_modulation_nodate}, \cite{gao_crosstlnet_nodate}, \cite{ding_data_2023}, \cite{xue_mlresnet_nodate}, \cite{riddhi_dual-stream_2024}, \cite{parmar_dual-stream_2023}, \cite{parmar_deep_2024}, \cite{chang_fast_2023}, \cite{luo_rlitnn_2024}, \cite{sun_novel_2023}, \cite{yang_irlnet_2021}, \cite{qu_enhancing_2024}, \cite{cheng_automatic_2024}, \cite{ying_convolutional_2023}, \cite{hou_signal_2023}, \cite{liu_deep_2024}, \cite{ma_automatic_2023} consistently provide the strongest balance of robustness and generalization. Architectures that combine CNNs with LSTMs, GRUs, or Transformers often achieve an average accuracy of 62\%-65\% on RML2016.10a, RML2016.10b, and RML2018.01a while maintaining moderate parameter sizes. These findings confirm hybridization as the most promising approach for feature-preserving AMC models suitable for real-world deployment. Additionally, AMC models integrating explicit denoising mechanisms demonstrate greater potential for balancing feature preservation with noise reduction, thereby achieving higher accuracy and generalization across diverse signal environments \cite{wei_adaptive_2023}, \cite{an_robust_2023}, \cite{suman_lightweight_2025}, \cite{an_efficient_2025}. In summary, CNNs dominate current practice; however, Transformers and hybrid designs demonstrate superior accuracy-to-complexity trade-offs and stronger cross-dataset performance, making them central to future advancements in AMC. Appendix A summarizes existing AMC models and reports performance results of CNN, RNN, Transformer, SNN, and Hybrid AMC models.\newline
The reviewed literature underscores substantial progress in AMC through innovations in neural architectures, data augmentation strategies, and feature extraction techniques. However, the three identified gaps, namely SNR sensitivity, computational complexity, and real-world validation, present persistent obstacles to translating academic advances into operational effectiveness. The first and most consequential gap is SNR sensitivity. While some recent architectures integrate denoising modules, no model consistently maintains high accuracy across both low and high SNR ranges. This limitation directly impacts the reliability of AMC in unpredictable field conditions. Addressing this gap would significantly enhance spectrum sensing, modulation recognition, and adaptive communications in dynamic environments. Lack of real-world training data raises serious concerns about model generalization. Synthetic datasets remain essential for benchmarking, but they must be supplemented with hardware-in-the-loop, semi-synthetic, or field-collected data to ensure robustness in real-world deployments \cite{zhang_deep_2022}. Without such validation, claims of operational readiness remain speculative.\newline 
Given these findings, the proposed study is both necessary and timely. This study will address critical shortcomings in the field by explicitly targeting SNR robustness while balancing complexity and performance. In addition, existing approaches treat I/Q coupling implicitly through convolution or complex-valued transformations. However, none explicitly frame this interaction as a structured, low-dimensional relational problem. This study addresses this gap by introducing a formal separation between instantaneous cross-channel interaction and temporal dynamics, enabling targeted modeling of each component.

\section{Problem Statement, Hypothesis, and Research Question}\label{sec1}
\subsection{Problem Statement}\label{subsec1}
The problem is that deep learning automatic modulation modes fail to perform consistently across low and high SNR conditions because existing feature-extraction methods suppress not only noise but also discriminative signal features.
\subsection{Hypothesis}\label{subsec2}
If a feature-preserving denoising solution is developed, then the root cause of losing discriminative features during feature extraction will be addressed.
\subsection{Research Question}\label{subsec3}
How can a feature-preserving denoising solution mitigate the loss of discriminative features during feature extraction in deep learning AMC models?

\section{Methods}\label{sec1}
\subsection{Proposed Model}\label{subsec1}
\begin{figure}[t]
\centering
\includegraphics[width=1.0\linewidth]{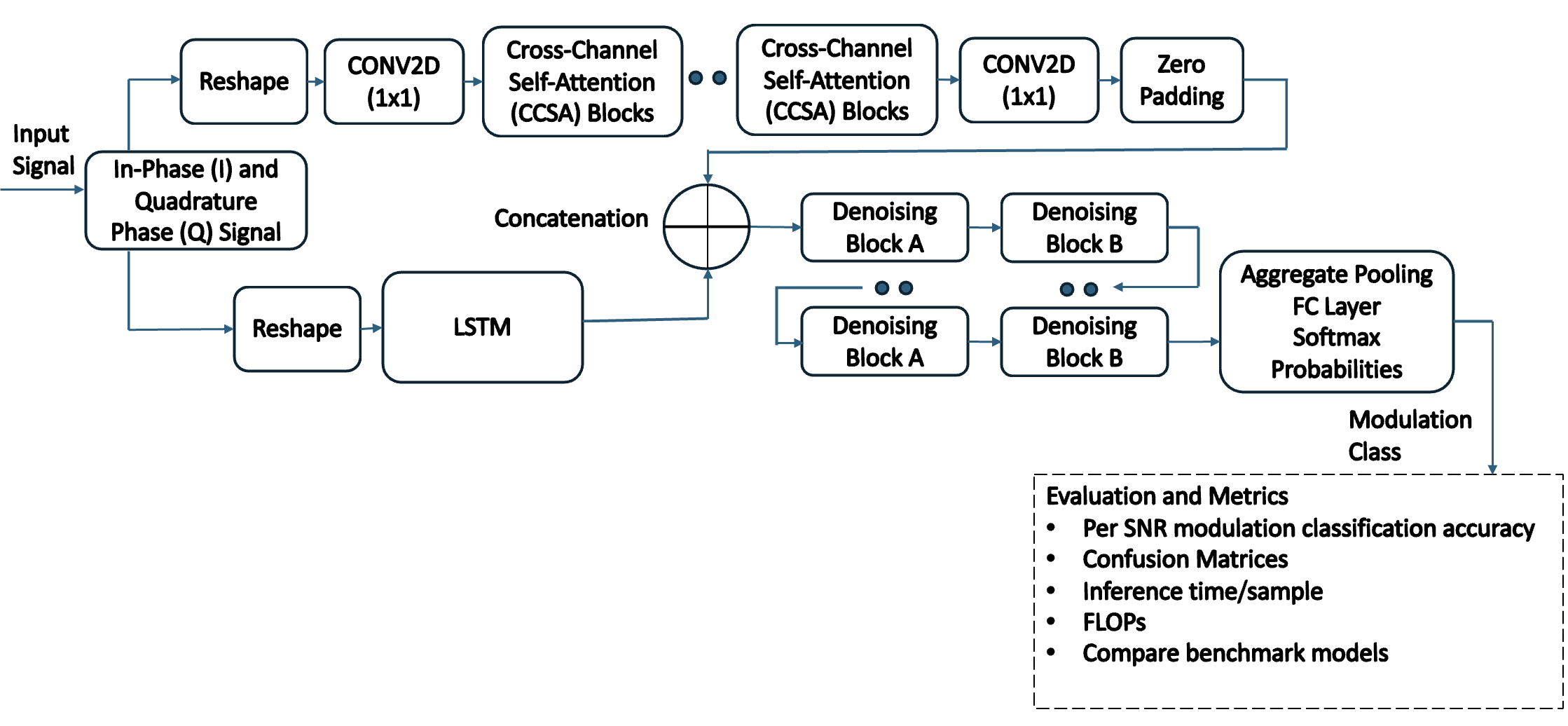}
\caption{Proposed model architecture}
\label{fig:image1}
\end{figure}

The architecture is guided by structural properties of complex baseband I/Q signals. Each sample is represented by coupled in-phase (I) and quadrature (Q) components that jointly encode amplitude and phase in a low-dimensional space. We therefore model I and Q as two token representations and apply cross-channel self-attention at each time step to learn their instantaneous dependencies. In this design, attention functions as a cross-channel mixing operator without requiring full temporal attention.\newline
Temporal dynamics and cross-channel interaction arising from different physical effects are modeled separately. Temporal structure reflects symbol transitions, while cross-channel structure reflects instantaneous complex projection. The model uses a temporal branch for sequence dynamics and a cross-channel attention branch for I/Q coupling, followed by feature fusion. Feature-preserving denoising is applied after feature fusion. Early denoising can remove low-amplitude yet discriminative structures at low SNR, whereas post-fusion adaptive noise suppression retains learned modulation features using data-driven thresholds.\newline
Hence, the proposed model architecture is a dual-branch neural network for AMC that explicitly separates temporal dynamics from cross-channel interactions between in-phase (I) and quadrature-phase (Q) components, and then reunifies them within a feature-preserving denoising backbone, as shown in Fig.~\ref{fig:image1}. One branch utilizes an LSTM to summarize sequence dynamics, while the other employs stacked cross-channel self-attention blocks to model instantaneous I–Q coupling at each time step. The fused representation then traverses a cascade of dual-path residual shrinkage blocks (DP-DRSN) as proposed by Suman and Qu \cite{suman_lightweight_2025} using garrote thresholding, which learn channel-wise, data-driven thresholds to attenuate noise while preserving discriminative coefficients, before undergoing global aggregation and softmax classification.

\subsection{Cross-Channel Self-Attention Block}\label{subsec2}
In wireless communication systems, raw baseband signals are typically represented as complex sequences with in-phase (I) and quadrature (Q) components. Efficient feature extraction from these two correlated streams is essential for tasks such as modulation classification, channel estimation, and interference detection. Traditional convolutional architectures often employ asymmetric filters to capture local temporal or frequency dependencies. However, such filters predominantly extract intra-channel features and are limited in modeling the inherent cross-channel dependencies between I and Q components. To address this gap, we propose a cross-channel self-attention mechanism that explicitly models the interaction between I and Q components at each time step. This approach leverages principles of multi-head self-attention (MHSA) from the Transformer architecture, reformulated for the special case of dual-channel I/Q signals.\newline
Let the input complex baseband signal be expressed as (1), where I(t) and Q(t) denote the in-phase and quadrature components, and T represents the signal length. The input is arranged as a tensor $X \in \mathbb{R}^{B \times T \times 2}$, where B is the batch size.
\begin{equation}
x(t) = I(t) + jQ(t), \quad t = 1,2,\ldots,T
\end{equation}
Each time step t is treated as a two-token sequence {I(t),Q(t)}. To map a scalar token into a higher-dimensional latent space with an embedding dimension d. A linear projection is implemented via a 1 x 1 convolution prior to the cross-channel self-attention block, as shown in (2), where $\mathbf{H} \in \mathbb{R}^{B \times T \times 2 \times d}$ where $\mathbf{H}$ represents the token embedding, $B$ denotes the batch size, $\mathbf{X}$ is the input tensor, $W_e$ is the weight matrix, and $b_e$ is the bias term.
\begin{equation}
\mathbf{H}_t = \mathbf{X}_t W_e + b_e,
\quad W_e \in \mathbb{R}^{1 \times d}
\end{equation}

For each time step $t$, self-attention across the two tokens $(I,Q)$ is applied.
Let $d_k$ denote the dimensionality of the queries and keys per head, and
$\mathbf{H}_t \in \mathbb{R}^{2 \times d}$ represent the token embedding at
time step $t$. The query ($M$), key ($K$), and value ($V$) matrices are
defined as

\begin{equation}
M = \mathbf{H}_t W_Q,\quad
K = \mathbf{H}_t W_K,\quad
V = \mathbf{H}_t W_V
\label{eq:qkv}
\end{equation}

where $W_Q, W_K, W_V \in \mathbb{R}^{d \times d_k}$ are learnable projection
matrices. The scaled dot-product attention is given by

\begin{equation}
\mathrm{Attention}(M,K,V) =
\mathrm{softmax}\left(\frac{M K^{\top}}{\sqrt{d_k}}\right)V
\label{eq:attention}
\end{equation}
Since each sequence has only two tokens, this explicitly models pairwise interaction between I and Q. For multi-head Multi-head self-attention (MHSA) is applied, where $h$ denotes the number of
attention heads. Each head computes attention separately with a reduced
dimension $d_k = d/h$. The outputs from all heads are concatenated, as shown
in~(5), where $W_o \in \mathbb{R}^{h d_k \times d}$ and $\oplus$ denotes
concatenation.

\begin{equation}
\mathrm{MHSA}(\mathbf{H}_t) =
\left( \bigoplus_{i=1}^{h}
\mathrm{Attention}(M_i,K_i,V_i) \right) W_o
\label{eq:mhsa}
\end{equation}

To further enhance representation power, each token is passed through a
position-wise feed-forward network (FFN) transformation, as shown in~(6),
where $\mathbf{z}$ represents the token embedding after MHSA,
$d_{ff}=4d$ is the hidden dimension of the FFN,
$W_1 \in \mathbb{R}^{d \times d_{ff}}$ is the first FFN weight matrix,
and $W_2 \in \mathbb{R}^{d_{ff} \times d}$ is the second FFN weight matrix.

\begin{equation}
\mathrm{FFN}(\mathbf{z}) =
\sigma(\mathbf{z} W_1 + b_1) W_2 + b_2
\label{eq:ffn}
\end{equation}

where $\sigma(\cdot)$ denotes the nonlinear GeLU activation function. Residual connections and layer normalization stabilize learning,
as shown in~(7) and~(8), where $\tilde{\mathbf{H}}_t$ denotes the
attention-updated feature produced by the multi-head self-attention
sublayer with residual connection and layer normalization.
The final block output $\mathbf{Z}_t$ represents the transformed
cross-channel features after the feed-forward sublayer, residual connections and layer
normalization are applied. The intermediate representation is given by

\begin{equation}
\tilde{\mathbf{H}}_t =
\mathrm{LayerNorm}\left(
\mathbf{H}_t + \mathrm{MHSA}(\mathbf{H}_t)
\right)
\label{eq:residual1}
\end{equation}

\begin{equation}
\mathbf{Z}_t =
\mathrm{LayerNorm}\left(
\tilde{\mathbf{H}}_t +
\mathrm{FFN}(\tilde{\mathbf{H}}_t)
\right)
\label{eq:residual2}
\end{equation}

The enhanced features from LayerNormalization are projected to $C$
channels using a $1\times1$ convolution after the cross-channel
self-attention block, yielding a feature tensor $F \in \mathbb{R}^{B \times T \times 2 \times C}$ as defined in~(9).

\begin{equation}
F_t = \mathbf{Z}_t W_p,
\quad
W_p \in \mathbb{R}^{d \times C}
\label{eq:projection}
\end{equation}

\begin{figure}[t]
\centering
\includegraphics[width=1.0\linewidth]{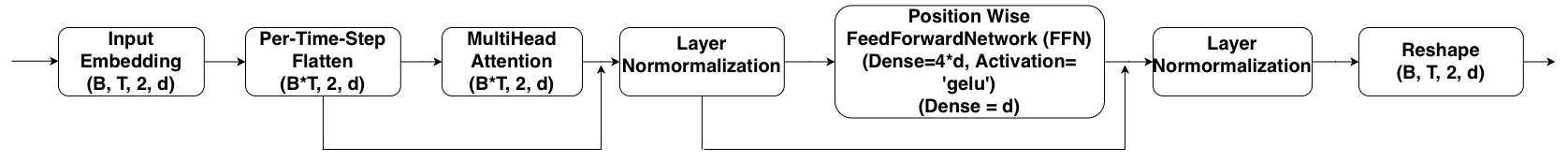}
\caption{Cross-channel self-attention block}
\label{fig:image2}
\end{figure}

The proposed cross-channel attention can be interpreted as a generalized complex interaction operator. Unlike complex multiplication, which imposes fixed algebraic coupling, or convolution, which implicitly mixes channels through learned kernels, attention enables data-dependent, directional interactions between $I$ and $Q$ components. This mechanism allows the model to adaptively emphasize latent relationships associated with amplitude and phase variations under different signal conditions, as illustrated in Fig.~\ref{fig:image2}.
In addition, restricting attention to two tokens mimics complex
multiplication, where $I$ and $Q$ interact to define amplitude and phase, while the computational complexity per time
step is $\mathcal{O}(2^{2} d)$, which is significantly more efficient
than full-sequence attention with complexity $\mathcal{O}(T^{2} d)$.

\subsection{Denoising Block}\label{subsec3}
The theory of denoising complex, noisy signals is rooted in the continuous wavelet transform (CWT). CWT has been widely applied in image compression, denoising, and signal processing \cite{chen_wavelet-based_2013}. Its ability to represent signals simultaneously in time and frequency domains makes it particularly valuable in radio signal propagation, where received signals are inherently non-stationary. Unlike the Fourier transform, which loses temporal localization, CWT provides a convolution-based time–frequency representation that enables dynamic threshold selection for effective noise suppression. Donoho’s \cite{donoho_-noising_1995} seminal work introduced the idea of thresholding or shrinkage functions in the wavelet transform domain, which became the foundation for modern denoising techniques. The central principle is that filters must preserve essential structures while attenuating stochastic noise, a process that has traditionally required domain expertise in filter design. Deep learning (DL) simplifies this by learning optimal filters directly through gradient descent, allowing for adaptive thresholding strategies.\newline

In this study, the dual-path deep residual shrinkage network (DP-DRSN)
framework employs garrote-thresholding denoising. The DP-DRSN proposed
by Suman and Qu~\cite{suman_lightweight_2025} is used for denoising by
combining global average pooling (GAP) and global max pooling (GMP)
to generate adaptive thresholds. GAP captures contextual information,
whereas GMP emphasizes high-amplitude features associated with
transient noise.

The threshold is calculated as a learnable convex combination of the
GAP-derived coefficient $\alpha$ and GMP-derived coefficient $\beta$,
weighted by $\gamma \in [0,1]$ and scaled by a self-learned factor
$\kappa$, as shown in~(10). This design enables dynamic adjustment
of denoising sensitivity during training, ensuring robustness in
diverse noisy environments.

\begin{equation}
\mathrm{Threshold} =
\kappa \left(\gamma \alpha + (1-\gamma)\beta \right)
\label{eq:threshold}
\end{equation}

\begin{figure}[t]
\centering
\includegraphics[width=1.0\linewidth]{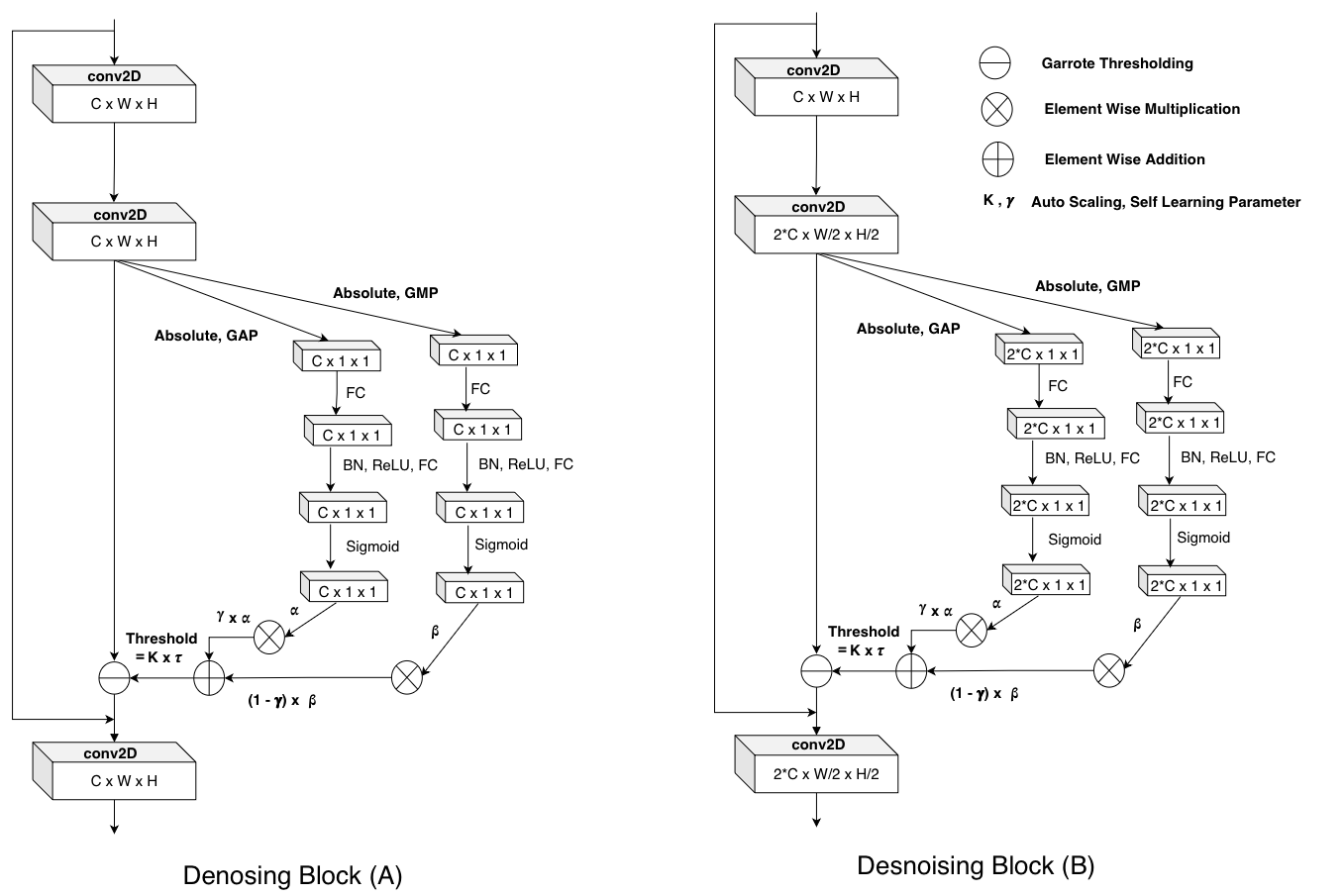}
\caption{Denoising block A and B}
\label{fig:image3}
\end{figure}

The structure of DP-DRSN begins with a subnetwork that takes an intermediate feature map as input. To generate adaptive thresholds, the model applies two pooling operations to the absolute values of the GAP and GMP features. GAP summarizes the overall distribution of signal features, providing contextual information across the entire feature map. GMP, by contrast, highlights the strongest individual responses, which often represent sudden peaks or bursts attributable to noise or interference. These pooled results are then passed into two fully connected layers with batch normalization and ReLU activation to produce scaling values. Finally, a sigmoid function ensures that the learned scaling parameters remain within a normalized range between 0 and 1. In this way, GAP and GMP provide complementary information that the network combines to set adaptive denoising thresholds.\newline
The denoising blocks based on DP-DRSN refine this process to achieve
robust feature preservation. As illustrated in Fig.~\ref{fig:image3}, Denoising
Block~A takes an input feature map of size
$C \times W \times H$. The block computes absolute values and applies
global average pooling (GAP) and global max pooling (GMP), each
followed by two fully connected layers with batch normalization,
ReLU, and sigmoid activation to generate scaling coefficients
$\alpha$ and $\beta$. These coefficients are then combined using a
self-learnable parameter $\gamma$, producing a convex combination of
the pooled statistics. A second learnable parameter $\kappa$ scales the combined threshold value, which is applied to the input feature map through garrote thresholding. This enables the network to adaptively suppress noise while retaining the signal's discriminative components.\newline
Denoising Block B extends this process by adding an additional downsampling step. Specifically, it applies strided operations with stride = 2, which reduces the spatial dimensions W and H by half while doubling the number of channels from C to 2C. This modification enables multi-scale feature learning by allowing the block to capture higher-level representations with increased channel capacity, while still applying the same adaptive thresholding mechanism as in Block A. By combining dimension reduction and channel expansion with the thresholding strategy, Denoising Block B provides a balance between feature compression and discriminative enhancement, ensuring that denoising occurs effectively across multiple levels of abstraction.
Together, Denoising Block A and Denoising Block B form the backbone of the adaptive denoising process in this study’s AMC model, providing a flexible, multi-scale mechanism for noise suppression and feature preservation across variable SNR conditions.

\subsection{Garrote Thresholding}\label{subsec4}

\begin{figure}[t]
\centering
\includegraphics[width=0.85\linewidth]{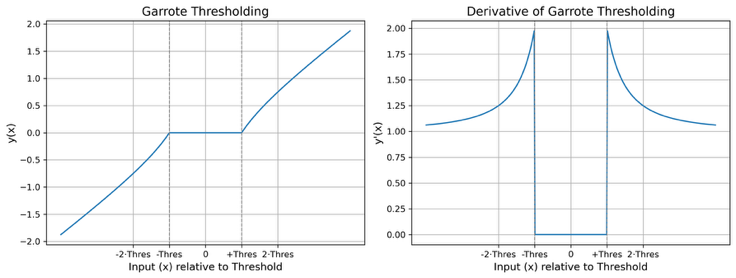}
\caption{Garrote thresholding and its derivative}
\label{fig:image4}
\end{figure}

In garrote thresholding, coefficients below the threshold are set to zero,
while coefficients above the threshold undergo nonlinear attenuation.
This formulation preserves large-magnitude coefficients while penalizing
small amplitudes, thereby avoiding the over-smoothing observed with soft
thresholding. Its derivative provides smooth, piecewise gradients that
stabilize backpropagation while maintaining sensitivity to informative
features.Suman and Qu~\cite{suman_lightweight_2025} applied garrote thresholding
to I/Q signals using an LSTM--CNN hybrid architecture, showing that
garrote thresholding effectively denoises signals while retaining
discriminative features. Equations~(11) and~(12) define garrote
thresholding and its derivative, respectively. Fig.~\ref{fig:image4} illustrates
garrote thresholding and its derivative. Note that $1 \times 10^{-6}$ is a small constant added to the denominator
to prevent numerical instability and ensure smooth behavior during
gradient computation.

\begin{equation}
y_i =
\begin{cases}
0, & |x| < \mathrm{Threshold} \\
x - \dfrac{\mathrm{Threshold}^2}{x + 10^{-6}}, & |x| \geq \mathrm{Threshold}
\end{cases}
\label{eq:garrote}
\end{equation}

\begin{equation}
y_i' =
\begin{cases}
0, & |x| < \mathrm{Threshold} \\
1 + \dfrac{\mathrm{Threshold}^2}{(x + 10^{-6})^2}, & |x| \geq \mathrm{Threshold}
\end{cases}
\label{eq:garrote_derivative}
\end{equation}

\subsection{Proposed Model Architecture}\label{subsec5}

\begin{figure}[t]
\centering
\includegraphics[width=1.0\linewidth]{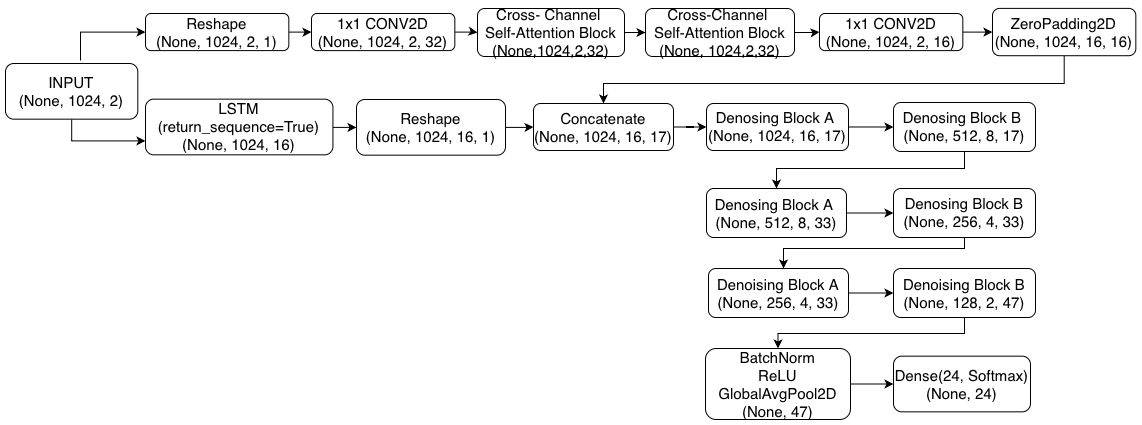}
\caption{Proposed model architecture with tensor shapes}
\label{fig:image5}
\end{figure}

Fig.~\ref{fig:image5} shows the full architecture of the proposed model with tensor shapes.
The proposed model ingests complex baseband wireless signals, in which each symbol is represented by its in-phase (I) and quadrature (Q) components. Each signal sequence is formatted as a tensor of the shape (B, T, 2), with B denoting batch size (128), T the sequence length (1024), and 2 corresponding to I and Q channels. The model has dual branches designed to extract complementary features.  The temporal branch extracts a temporal representation and then fuses with the multi-layer cross-channel self-attention branch.\newline
To match the LSTM branch with dimensions
$B \times T \times \mathrm{Units} \times 1$, zero padding is applied
along the width dimension of $F_t$ from~(8), expanding it from 2 to
$\mathrm{Units}$ and resulting in $F_{\text{padded}} \in \mathbb{R}^{B \times T \times \mathrm{Units} \times C}$. Parallel to the attention branch, an LSTM extracts temporal
dependencies, producing the representation
$\mathbf{H}_{\mathrm{LSTM}}$, as defined in~(13).
\begin{equation}
\mathbf{H}_{\mathrm{LSTM}} =
\mathrm{LSTM}(\mathbf{X}),
\quad
\mathbf{H}_{\mathrm{LSTM}} \in
\mathbb{R}^{B \times T \times \mathrm{Units}}
\label{eq:lstm}
\end{equation}

After reshaping $\mathbf{H}_{\mathrm{LSTM}}$ to $\mathbf{H}_{\mathrm{LSTM}}' \in \mathbb{R}^{B \times T \times \mathrm{Units} \times 1}$, the final fused feature representation is defined as
$\mathbf{F}_{\mathrm{fusion}}$, as shown in~(14).

\begin{equation}
\mathbf{F}_{\mathrm{fusion}} =
\mathrm{Concat}
\left(
F_{\text{padded}},
\mathbf{H}_{\mathrm{LSTM}}'
\right),
\quad
\mathbf{F}_{\mathrm{fusion}} \in
\mathbb{R}^{B \times T \times \mathrm{Units} \times (C+1)}
\label{eq:fusion}
\end{equation}
The outputs of the two branches are fused through a sequence of denoising blocks A and B. Each denoising block applied convolutional feature extraction followed by adaptive thresholding. Threshold values are dynamically estimated via GAP and GMP, followed by fully connected layers with sigmoid activations to compute scaling coefficients. Learnable parameters adjust these thresholds adaptively during model training, enabling the model to respond to variable SNR conditions.\newline
After denoising, the outputs are passed through a batch normalization layer and activated with a ReLU activation. The GAP operation aggregated the spatiotemporal information into a compact representation vector, significantly reducing dimensionality while retaining discriminative features. The pooled representation is fed into a fully connected layer with softmax units, corresponding to the modulation classes. This produces the final probability distribution over possible modulation types. The model is trained using the categorical cross-entropy loss function, which compares the prediction softmax probabilities with the true modulation labels. Optimization is carried out using the Adam optimizer with a learning rate empirically tuned for convergence. L2 regularization is applied to convolutional and fully connected layers to reduce overfitting and stabilize weight updates. The architecture reflects a design principle that involves separate modeling of temporal dynamics and instantaneous I/Q interaction, followed by adaptive denoising to preserve discriminative features.

\section{Results}\label{sec1}
\subsection{Dataset}\label{subsec1}
\begin{figure}[t]
\centering
\includegraphics[width=0.85\linewidth]{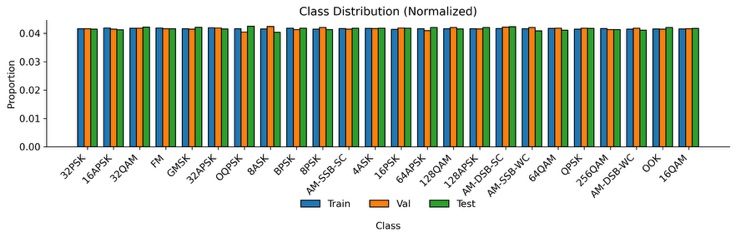}
\caption{Normalized modulation class distribution for train, validation, and test samples}
\label{fig:image6}
\end{figure}

\begin{figure}[t]
\centering
\includegraphics[width=0.85\linewidth]{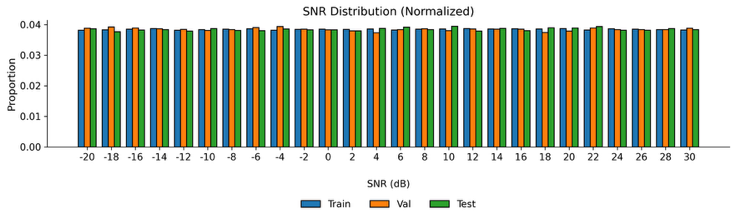}
\caption{Normalized SNR level distribution for train, validation, and test samples}
\label{fig:image7}
\end{figure}

The RML2018.01a dataset is used in this study, which simulates diverse
conditions under which automatic modulation classification (AMC)
models must operate. The study sample was drawn from the RML2018.01a
dataset, consisting of 2,555,904 samples, with 4096 samples for each
modulation and signal-to-noise ratio (SNR) level (i.e.,
$24 \text{ modulations} \times 26 \text{ SNR levels} \times 4096$),
totaling 2,555,904 samples across 24 modulation schemes, including
32PSK, 16APSK, 32QAM, FM, GMSK, 32APSK, OQPSK, 8ASK, BPSK, 8PSK,
AM--SSB--SC, 4ASK, 16PSK, 64APSK, 128QAM, 128APSK, AM--DSB--SC,
AM--SSB--WC, 64QAM, QPSK, 256QAM, AM--DSB--WC, and OOK, with SNR
values ranging from $-20\,\mathrm{dB}$ to $+30\,\mathrm{dB}$.
A stratified random sampling method was used to ensure uniform
representation of all modulation classes and SNR levels. The dataset
was split into 1,000,000 training samples, 200,000 validation samples,
and 200,000 test samples. Fig.~\ref{fig:image6} and Fig.~\ref{fig:image7} show the normalized
proportions for the training, validation, and test sets.

\subsection{Experimental Setup}\label{subsec2}
The experimental setup included TensorFlow~3.9.19 and Keras~2.10.1
frameworks for training and evaluating deep learning models on a
Windows~11 workstation equipped with an Intel Core~i9 processor,
64~GB of RAM, and an NVIDIA GeForce RTX~4080 GPU. Data collected
during evaluation included classification accuracy across all
modulation classes and signal-to-noise ratio (SNR) levels.
Reliability measures were conducted to confirm reproducibility
across models using fixed random seeds for stratified training,
validation, and test data generation. The dataset was partitioned into stratified training
($1{,}000{,}000$ samples), validation ($200{,}000$ samples),
and testing ($200{,}000$ samples) subsets across SNR and
modulation types.

The training process used a batch size of $128$, an initial
learning rate of $0.001$, and a minimum learning rate of
$10^{-8}$, with Adam optimizer parameters
$\beta_1 = 0.9$ and $\beta_2 = 0.999$.
Each model was trained for $200$ epochs, and the learning
rate was reduced by a factor of two if the validation loss
did not improve after five epochs. Using the \texttt{Keras EarlyStopping} procedure, training
was terminated if the validation loss did not decrease after
30 epochs. Classification accuracy at each SNR level,
inference latency per sample, floating-point operations
(FLOPs), and model parameter counts were collected for
each model. All model evaluations were conducted using
garrote thresholding as the denoising mechanism.

\subsection{Testing and Traning Results}\label{subsec3}

\begin{figure}[t]
\centering
\includegraphics[width=0.85\linewidth]{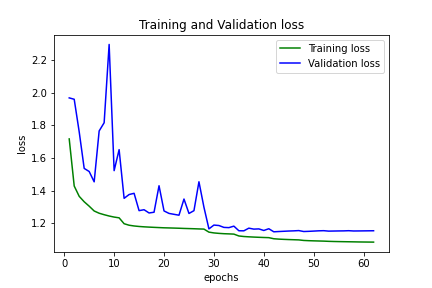}
\caption{Proposed model training and validation loss}
\label{fig:image8}
\end{figure}

\begin{figure}[t]
\centering
\includegraphics[width=0.85\linewidth]{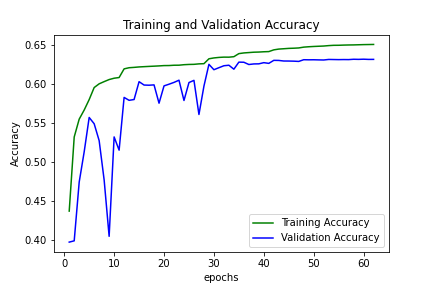}
\caption{Proposed model training and validation accuracy}
\label{fig:image9}
\end{figure}

\begin{figure}[t]
\centering
\includegraphics[width=0.85\linewidth]{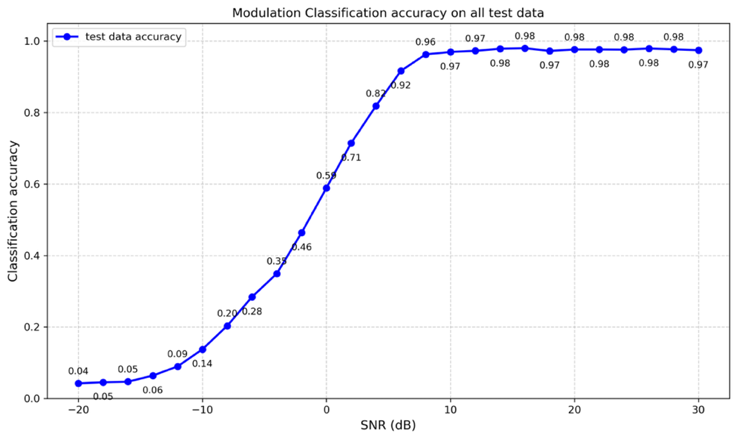}
\caption{Proposed model modulation classification accuracy at each SNR level}
\label{fig:image10}
\end{figure}

The training and validation loss and accuracy curves of the proposed
model are shown in Fig.~\ref{fig:image8} and Fig.~\ref{fig:image9}. Fig.~\ref{fig:image10} further illustrates the
modulation classification accuracy of the proposed model at each
signal-to-noise ratio (SNR) level on the test dataset. The observed performance gains at low SNR indicate that denoising plays a critical role in preserving class-separating features. The increasing accuracy across different SNR levels indicates that maintaining features, not just reducing noise, is key to performance improvements.
In addition, a confusion matrix was generated at each SNR level. Appendix~B
presents the confusion matrices at SNR levels of
$-4\,\mathrm{dB}$, $0\,\mathrm{dB}$, $4\,\mathrm{dB}$, and
$30\,\mathrm{dB}$.

To assess the stability of the proposed automatic modulation
classification (AMC) model, five-fold cross-validation was conducted
on $1.2$ million samples. For each fold,
$1.0$ million samples were used for training and
200k samples for validation, with the validation
subset systematically rotated across the $1.2$ million samples.
Model performance was evaluated using overall accuracy,
macro-precision, macro-recall, and macro-F1 metrics.
Aggregate performance is reported as the mean and sample standard
deviation across folds, as shown in Table~\ref{tab:cv_results}.

Across the five folds, the proposed model achieved an average
overall classification accuracy of
$62.6\% \pm 0.5\%$, with macro-precision of
$65.8\% \pm 1.2\%$, macro-recall of
$62.6\% \pm 0.5\%$, and macro-F1 score of
$62.9\% \pm 0.5\%$.
The consistently low standard deviation observed across all metrics
indicates that the model performance is stable with respect to
variations in training and validation splits, suggesting limited
sensitivity to specific data partitions.

\begin{table}[t]
\centering
\caption{Cross-validation results}
\label{tab:cv_results}
\begin{tabular}{c c c c c}
\toprule
Fold & Overall Accuracy (\%) & Macro-Precision (\%) & Macro-Recall (\%) & Macro-F1 (\%) \\
\midrule
0 & 63.21 & 67.13 & 63.73 & 63.73 \\
1 & 62.61 & 64.38 & 62.65 & 62.65 \\
2 & 62.38 & 65.48 & 62.67 & 62.67 \\
3 & 62.88 & 66.70 & 63.32 & 63.32 \\
4 & 61.88 & 65.15 & 62.26 & 62.26 \\
\midrule
Mean $\pm$ Std & 
$62.6 \pm 0.5$ & 
$65.8 \pm 1.2$ & 
$62.6 \pm 0.5$ & 
$62.9 \pm 0.5$ \\
\bottomrule
\end{tabular}
\end{table}

\begin{figure}[t]
\centering
\includegraphics[width=0.85\linewidth]{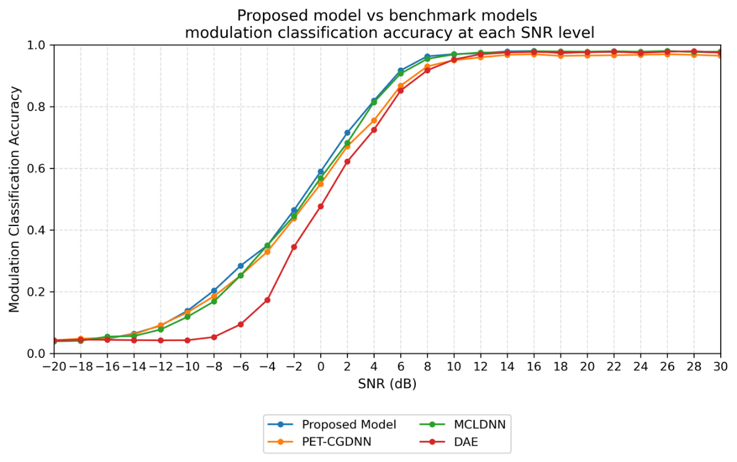}
\caption{Proposed model vs benchmark models modulation classification accuracy at each SNR level}
\label{fig:image11}
\end{figure}

The close alignment between macro-recall and overall accuracy further indicated that classification performance is balanced across modulation classes. In addition, the modest variation in macro-precision reflects consistent modulation class-wise decision boundaries across folds, despite the presence of varying channel impairments and SNR conditions inherent in the dataset
In addition, the proposed model is compared with benchmark models to further evaluate its effectiveness under low noise conditions, as shown in Fig.~\ref{fig:image11}. Benchmark models were trained and tested under identical experimental conditions to those of the proposed model. The proposed model exhibits consistent improvement in modulation accuracy across low to moderate SNR conditions from -8dB to 2dB. The proposed model yields a larger modulation classification accuracy gain, approaching and surpassing 0 dB, compared to the benchmark models MCLDNN \cite{chang_multitask-learning-based_2022}, PET-CGDNN \cite{zhang_efficient_2021}, and DAE \cite{ke_real-time_2022}. Under low-noise conditions, the proposed model is consistently robust, achieving higher accuracies from -8 dB to +2dB with an average gain of 3\% over PET-CGDNN, 2.3\% over MCLDNN, and 14\% over DAE. Appendix C provides raw data on average classification accuracy at each SNR level for the proposed and benchmark models used in the study.

\section{Discussion}\label{sec1}
\subsection{Ablation Study}\label{subsec1}
A systematic ablation study was conducted to analyze fourteen model architectures and assess the influence of feature-preserving denoising on classification performance, as shown in Table~\ref{tab:ablation}. The results of the model performance, average classification accuracy, maximum classification accuracy, and inference time/sample (ms), together with model training parameters size and FLOPs.  All models used garrote thresholding as a denoising approach, and all shared a uniform experimental setup with a fixed random seed and an identical dataset with samples distributed evenly across all SNR levels and modulation classes, ensuring that no class or SNR-specific bias affected the results.

\begin{table*}[t]
\centering
\caption{Ablation study of the proposed model configurations.}
\label{tab:ablation}

\resizebox{\textwidth}{!}{%
\begin{tabular}{cccccccccccccc}
\toprule
Model & LSTM & $d$ & Heads & CCSA & Den. & Den.& Thres. & Params & FLOPs & Avg. Acc. & Max Acc. & Inference time/ \\
No. & Units &  &  & Blocks & Block A & Block B &  &  & (M) & (\%) & (\%) & sample (ms) \\
\midrule
1  & 16 & 32 & 2 & 1 & 3 & 3 & Garrote & 165,580 & 234  & 62.79 & 97.88 & 0.53 \\
2  & 16 & 32 & 2 & 2 & 3 & 3 & Garrote & 178,284 & 270  & 63.51 & 98.00 & 0.68 \\
3  & 16 & 32 & 2 & 3 & 3 & 3 & Garrote & 190,988 & 306  & 63.47 & 98.13 & 0.79 \\
4  & 16 & 32 & 2 & 2 & 2 & 2 & Garrote & 85,354  & 232  & 62.77 & 97.64 & 0.63 \\
5  & 16 & 32 & 2 & 2 & 1 & 1 & Garrote & 40,878  & 161  & 57.96 & 92.74 & 0.58 \\
6  & 16 & 32 & 2 & 2 & -- & -- & Garrote & 27,682  & 74   & 40.37 & 72.81 & 0.44 \\
7  & -- & 32 & 2 & 2 & -- & -- & Garrote & 26,440  & 74   & 38.91 & 73.68 & 0.30 \\
8  & 16 & -- & -- & -- & -- & -- & Garrote & 1,266   & 0.02 & 16.41 & 22.57 & 0.19 \\
9  & 16 & 32 & 2 & 3 & -- & -- & Garrote & 40,386  & 110  & 40.94 & 73.06 & 0.64 \\
10 & 16 & 64 & 2 & 3 & 3 & 3 & Garrote & 303,404 & 617  & 63.19 & 97.85 & 1.36 \\
11 & 32 & 32 & 2 & 2 & 2 & 2 & Garrote & 173,664 & 1030 & 63.52 & 97.97 & 1.48 \\
12 & 32 & 32 & 2 & 3 & 3 & 3 & Garrote & 186,368 & 1060 & 63.31 & 98.07 & 1.55 \\
13 & 32 & 32 & 2 & 2 & 4 & 4 & Garrote & 341,380 & 1100 & 63.34 & 98.00 & 1.45 \\
14 & 16 & 32 & 2 & 2 & 4 & 4 & Garrote & 346,000 & 287  & 63.10 & 97.92 & 0.72 \\
\bottomrule
\end{tabular}%
}

\end{table*}

\begin{figure}[t]
\centering
\includegraphics[width=0.85\linewidth]{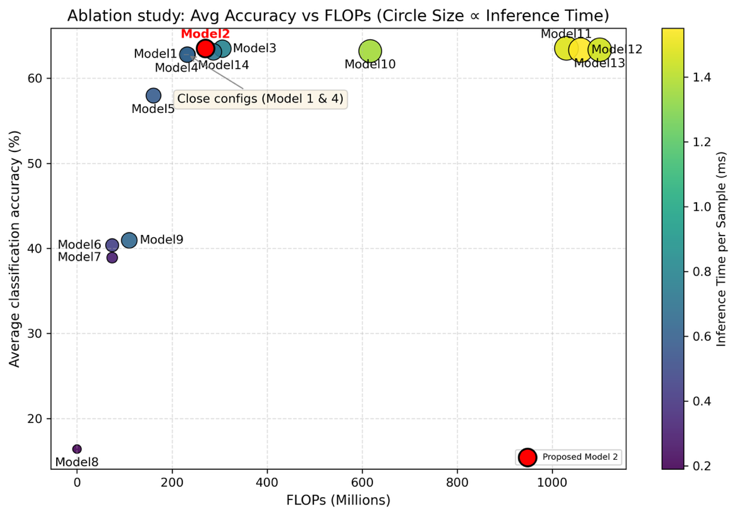}
\caption{Ablation study: Average modulation classification accuracy vs. FLOP; color by inference time/sample (ms)}
\label{fig:image12}
\end{figure}

\begin{figure}[t]
\centering
\includegraphics[width=0.85\linewidth]{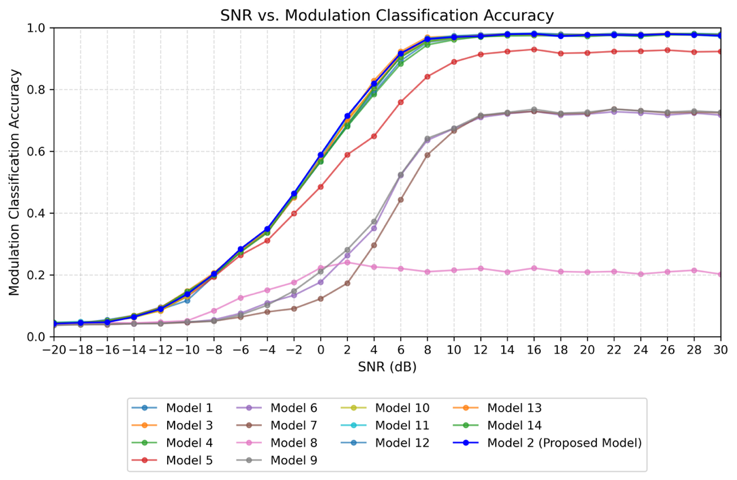}
\caption{Ablation study: average modulation classification accuracy per SNR level}
\label{fig:image13}
\end{figure}

Each model variant modified one or more of four key design factors: (i) the number of cross-channel self-attention (CCSA) blocks, (ii) the depth of denoising block A and B, (iii) the embedding dimension (d) for multi-head attention, and (iv) the number of LSTM units in the temporal feature extraction stage. The experiments aimed to identify configurations that preserve discriminative features, resulting in high classification accuracy under noisy conditions while minimizing computational load and inference time/sample.\newline
\underline{Effect of CCSA Depth}: Models 1, 2, and 3 isolated the impact of varying numbers of CCSA blocks while maintaining identical LSTM, embedding, and denoising configurations. Results revealed that increasing CCSA depth from 1 to 2 blocks improved the average classification accuracy from 62.79\% to 63.51\%, but further deepening the three CCSA blocks yielded negligible improvement to 63.47\%, while increasing the computational cost by 36M FLOPs and the inference time by 0.11ms per sample. This demonstrated diminishing return beyond 2 CCSA blocks, suggesting that CCSA=2 captures sufficient cross-feature interaction without over-parameterization.
\underline{Effects of Denoising Block Depth}: Next models 2, 4, 5, 6, and 14 examined the impact of reducing or expanding denoising depth while fixing CCSA=2. Performance degraded consistently as the number of denoising layers decreased. The average accuracy dropped from 63.51\% with 3 denoising blocks (A and B) to 62.77\% with 2 denoising blocks (A and B), then to 57.96\% with 1 denoising block (A and B), and finally to 40.37\% when denoising was removed altogether. Model 14, which increases denoising depth to four denoising blocks A and B, does not improve performance relative to model 2 (63.10\% vs 62.51\%) despite a substantial increase in model training parameters (from 178k to 346k). These findings highlight that feature-preserving denoising is necessary to preserve the discriminative structure of noisy modulated signals. The 3 denoising block A and B configuration provides the best average classification accuracy. This indicates that the deeper the denoising blocks, the better the feature preservation, resulting in higher classification accuracy, though at a high computational cost.
Models 13 and 14 examined whether extending the denoising depth further enhanced recovery at high noise levels. In Model 14, although increasing from 2 denoising blocks A and B to 4 denoising blocks A and B compared to Model 11, the average accuracy slightly decreased to 63.34\% versus 63.52\%, while the number of parameters nearly doubled (341 K vs 173 K). In addition, Model 14 increased the denoising blocks A and B from 3 to 4 compared to Model 2. The average accuracy dropped to 63.10\% from 63.51\%, while the number of parameters doubled (346k vs 178k). Therefore, deeper denoising beyond 3 denoising blocks A and B adds computational complexity without a clear performance improvement.\newline
\underline{Effects of Removing Denoisers and Scaling CCSA}: Comparing Model 6, with 2 CCSA blocks and no denoising blocks, and Model 9, with 3 CCSA blocks and no denoising blocks, shows that an even deeper attention network cannot compensate for the absence of denoising. Average classification accuracy improved by only 0.6 percentage points, from 40.37\% to 40.94\%, confirming that attention alone cannot disentangle signal and noise features without a denoising stage. Models 6 and 7 provide further insight into architectures without denoising. Model 6 retains an LSTM with 2 CCSA blocks but no denoiser and achieves 40.37\% average accuracy. Model 7 removes both denoising and the LSTM, leaving only attention-based feature extraction, and accuracy falls slightly further to 38.91\%. This difference indicates that while LSTM contributes to temporal structure learning, attention-only architectures cannot compensate for the absence of denoising, even with increased depth. Increasing CCSA or widening attention cannot compensate for the loss of feature-preserving denoising capacity.\newline
\underline{Influence of Embedding Dimension (d)}: Model 10 doubled the embedding dimension from 32 to 64 while maintaining identical attention and denoising structure as Model 3. This modification increased FLOPs from 306 M to 617 M and inference latency from 0.79 ms to 1.36 ms, an over 70\% increase in computational load, yet slightly decreased average accuracy from 63.47\% to 63.19\% compared to Model 3. The larger embedding evidently expanded representational capacity without corresponding discriminative gain, indicating that (d=32) provides sufficient embedding richness for modulation classification.\newline
\underline{Influence of LSTM Width}: Models 4 and 11, both with CCSA=2 and 2 denoiser blocks A and B and d=32, compared the impact of doubling LSTM units from 16 to 32. The wider models achieved 63.52\% average accuracy versus 62.77\% but the cost increased substantially, nearly 4x FLOPs (from 232M to 1030M) and 0.85ms inference time/sample. A similar comparison between Models 3 and 12 (CCSA=3 with 3 denoising blocks A and B) showed the same trend; accuracy stagnated or slightly declined as the parameter ballooned. The pattern suggests that LSTM widening beyond 16 units offers limited marginal gain and may overfit temporal dependencies rather than enhancing robustness. Model 8 isolates the effect of the LSTM, serving as a lower bound on the model design space, and verifies that ultra-lightweight architectures sacrifice discriminative capacity.\newline
In summary, the ablation results collectively indicated that cross-channel self-attention depth and denoising depth are the primary factors influencing model performance. The ablation study established three design principles to maximize classification accuracy at the lowest cost: use CCSA=2, as deeper attention offers minimal accuracy gains but increases computational costs, and maintain robust denoising with 3 denoising blocks in the A and B configuration, which is key to achieving high accuracy; shallower or absent denoisers perform significantly worse. Lastly, avoid LSTM widening and large embedding dimensions, as both inflate computation costs without improving classification accuracy.\newline
Ablation study reveals that denosing depth is the key factor in enhancing low to medium SNR robustness, moderate CCSA effectively captured I/Q dependency, increasing tempral model capacity offeres limited improvements, and feature-preserving denoising is crucial for effective attention-based AMC. Based on these observations, Model 2, the proposed model, achieves the best average and maximum classification accuracy with the lowest computational burden, making it the most efficient option. Fig.~\ref{fig:image12} presents the trade-off between average classification accuracy, computational cost (measured in FLOPs), and inference time/sample (ms), illustrating how accuracy initially improves as model complexity increases, then plateaus beyond the mid-range from 232M to 306M FLOPs. Models 11, 12, and 13 show only marginal gains in accuracy relative to substantially higher-cost configurations, indicating diminishing returns in performance despite the increased computational burden. This flattening trend underscores that architectural choices, rather than sheer computational scale, drive the most meaningful improvements in accuracy. Fig.~\ref{fig:image13} compares the performance of Model 1 to Model 14 across different SNR levels.

\subsection{Limitation of Study Findings}\label{subsec2}
The study’s findings should be interpreted in light of several limitations related to data provenance, external validity, computational constraint, and comparative scope. All experiments were conducted on a standardized, publicly available dataset under controlled simulation settings. No live over-the-air data was collected, and no field deployment was executed. As a result, the results might not fully represent real-world conditions in operational radio settings, such as non-stationary interference, radio receiver’s front-end non-linearities, oscillator drift, spectrum congestion, or hardware-induced I/Q imbalance. This limits the generalizability and transferability of the results beyond hte simulated channels represeted in the dataset.\newline
Model training and inference profiling were executed on a single GPU workstation (Windows OS). This constrained the breadth of hyperparameter exploration and the depth/width of model architectures that could be practically evaluated. Direct comparison with prior AMC models was restricted to a small set of representative benchmark models. These baselines were re-implemented and re-evaluated within the same software/hardware environment to promote fairness. However, comprehensive external benchmarking across the full diversity of published AMC models was not feasible. As a result, the relative standing of the proposed model should be viewed as a promising method in the most challenging low-SNR conditions for future field testing.\newline
Despite stratified sampling across modulation classes and SNR levels and careful control of evaluation protocols, deep learning training remains sensitive to weight initialization, batch ordering, and optimizer dynamics. Fixed seeds and repeated trials mitigate but do not eliminate variance in outcomes. Additionally, while the chosen dataset is widely used and curated, any label noise or mismatch between simulated and real channel statistics could affect performance estimates. Together, these factors introduce uncertainty into the point estimates of classification accuracy and inference time.\newline
The study reinforces internal validity via: (a) uniform pre-processing and splits across of training/validation/test across all model variants, (b) identical optimization, training,  and early stopping criteria, (c) stratified evaluation at each SNR level, (d) efficiency-aware profiling (training parameters, FLOPs, inference time) reported alongside classification accuracy, and (e) consistent re-evaluation of baseline within the same environment. These controls strengthen the trustworthiness of the comparative claim within the studied design space. Nevertheless, the findings are best interpreted as evidence of the relative effectiveness and efficiency of feature-preserving denoising under controlled conditions.

\subsection{Recommendation for Future Study}\label{subsec3}
Building on the finding that denoising depth is the principal mediator of low-SNR reliability, future research should investigate innovative, feature-preserving denoising mechanisms, such as Quadrature Curve Thresholding (QCT) \cite{salimy_dynamic_2022}, that explicitly model I/Q dependencies during shrinkage. To strengthen external validity and address the limitation of simulated datasets, subsequent work should incorporate over-the-air data from real transmission environments characterized by non-stationary interference, hardware impairments, and congestion \cite{syed_deep_2023}.\newline
As shown in this study, given the observed compute-accuracy plateau and the practical imperative to improve spectral efficiency without excessive computational overhead, future research should reduce model complexity via pruning and quantization \cite{zhang_efficient_2021} while preserving low-SNR robustness. In addition, future work should test plateau behavior on real over-the-air datasets to distinguish architecture limits from dataset limits.\newline
Future studies should evaluate deployment feasibility across heterogeneous hardware platforms to ensure practical applicability across modern communication stacks common in 5G and future 6G infrastructure \cite{ansari_attention-enhanced_nodate}, including general-purpose CPUs, GPUs, and NPUs. To broaden generalization beyond a single corpus and propagation regime, future research should pursue transfer learning across multi-source datasets, including software-defined radio testbeds, UAV air-to-ground channels, and satellite links. Collectively, these future research directions roadmap translate this study's core inference into innovations for denoisers, validate performance on real-world data, develop efficient AMC models, explore heterogeneous hardware deployments, and ensure models' generalization across different domains.

\section{Conclusion}\label{sec1}
This research enhanced the understanding of how to establish robust AMC in low to moderate SNR environments. To address this limitation, we proposed an AMC architecture that integrates a cross-channel self-attention block to explicitly model I/Q interaction, and a feature-preserving dual-path residual shrinkage network employing adaptive garrote thresholding. The design formalizes I/Q interaction as a structured learning problem, establishing feature-representing denoising as the dominant factor in low to moderate SNR AMC, while effectively reducing noise without sacrificing modulation-separating features.
Extensive experiments on the RML2018.01a dataset demonstrated that the proposed model achieves stable and reproducible performance, yielding a cross-validated mean classification accuracy of 62.6\%, macro precision of 65.8\%, mean macro recall of 62.6\%, and mean macro F1 score of 62.9\% with low variance. Compared with PET-CGDNN, MCLDNN, and DAE baselines, the model consistently improved classification accuracy by 3\%, 2.3\%, and 14\%, respectively, across the SNR range from -8 dB to +2 dB. 
The study is limited by reliance on synthetic data and evaluation within a controlled computational environment, which constrains external validity. Future work will focus on validating the approach with over-the-air measurement, exploring alternative I/Q-aware denoising mechanisms, and reducing computational complexity through pruning and quantization to support real-world wireless systems.

\section*{List of Abbreviations}

\begin{longtable}{p{3cm} p{11cm}}
AMC & Automatic Modulation Classification \\
SNR & Signal-to-Noise Ratio \\
I/Q & In-phase / Quadrature \\
LSTM & Long Short-Term Memory \\
DP-DRSN & Dual-Path Deep Residual Shrinkage Network \\
CNN & Convolutional Neural Network \\
RNN & Recurrent Neural Network \\
SNN & Spiking Neural Network \\
DL & Deep Learning \\
MHSA & Multi-Head Self-Attention \\
FFN & Feed-Forward Network \\
GAP & Global Average Pooling \\
GMP & Global Max Pooling \\
CCSA & Cross-Channel Self-Attention \\
FLOPs & Floating-Point Operations \\
GPU & Graphics Processing Unit \\
RAM & Random Access Memory \\
OS & Operating System \\
QCT & Quadrature Curve Thresholding \\
DAE & Denoising Autoencoder \\
GRU & Gated Recurrent Unit \\
NPU & Neural Processing Unit \\
UAV & Unmanned Aerial Vehicle \\
PSK & Phase Shift Keying \\
APSK & Amplitude and Phase Shift Keying \\
QAM & Quadrature Amplitude Modulation \\
FM & Frequency Modulation \\
GMSK & Gaussian Minimum Shift Keying \\
OQPSK & Offset Quadrature Phase Shift Keying \\
ASK & Amplitude Shift Keying \\
BPSK & Binary Phase Shift Keying \\
QPSK & Quadrature Phase Shift Keying \\
OOK & On-Off Keying \\
AM-SSB-SC & Amplitude Modulation, Single Sideband, Suppressed Carrier \\
AM-DSB-SC & Amplitude Modulation, Double Sideband, Suppressed Carrier \\
AM-SSB-WC & Amplitude Modulation, Single Sideband, With Carrier \\
AM-DSB-WC & Amplitude Modulation, Double Sideband, With Carrier \\
\end{longtable}

\section*{Declarations}

\subsection*{Availability of data and material}
The datasets used in this study are provided by DeepSig Inc. and are available under the Creative Commons Attribution--NonCommercial--ShareAlike 4.0 License (CC BY-NC-SA 4.0) at \url{https://www.deepsig.ai/datasets/} (accessed on 5 August 2025).

\subsection*{Competing interest}
The authors declare no competing financial or non-financial interests related to this work.

\subsection*{Funding}
No funding was received for this study.

\subsection*{Authors contribution}
Dr. Prakash Suman conducted conceptualization, methodology, formal analysis, visualization, and original draft preparation. Dr. Yanzhen Qu contributed to review, editing, and supervision. All authors approved the final manuscript and are accountable for its content.

\subsection*{Acknowledgements}
The authors used Python (3.9.19) and TensorFlow (2.10.1) for analysis and visualization. All outputs were reviewed and validated by the authors.

\clearpage

\begin{appendices}

\section{}\label{secA1}

\setlength\LTleft{-2cm}
\setlength\LTright{0cm}

\small
\begin{longtable}{@{}
p{3.0cm}  
p{1.9cm}  
p{2.5cm}  
p{0.9cm}  
p{2.6cm}  
p{1.2cm}  
p{0.8cm}  
p{1.1cm}  
p{1.1cm}  
@{}}
\caption{Summary of existing CNN, RNN, Transformer, SNN, Hybrid deep learning AMC models; Dataset A: RML2016.10a, B: RML2016.10b, C: RML2018.01a, and NS: Non-standard dataset}
\label{tab:review_comparison}\\

\toprule
Author & Model Name & Input Signal/Pre-processing & Den. & DL Architecture & Trainable Parameters & Dataset & Avg Accuracy & Max Accuracy \\
\midrule
\endfirsthead

\toprule
Author & Model Name & Input Signal / Pre-processing & Den. & DL Architecture & Trainable Parameters & Dataset & Avg Accuracy & Max Accuracy \\
\midrule
\endhead

\midrule
\multicolumn{9}{r}{\footnotesize Continued on next page}\\
\midrule
\endfoot

\bottomrule
\multicolumn{9}{p{\textwidth}}{\footnotesize
\textbf{Notes:} Den.\ = denoising ($\checkmark$ = used). A/P = amplitude/phase}\\
\endlastfoot

\multicolumn{9}{c}{\textbf{CNN Models}}\\
\midrule
Shaik and Kirthiga~\cite{shaik_automatic_2021} & DenseNet & -- & -- & CNN & 19k & C & 55\% & -- \\
Shen et al.~\cite{shen_multi-subsampling_2023} & MSSA & I/Q & -- & CNN & 36k--218k & C & 55.25\%--60.90\% & -- \\
Li et al.~\cite{li_lightweight_2023} & LightMFFS & I/Q, A/P & -- & CNN & 95k & A & 63.44\% & -- \\
 &  &  &  &  & 95k & B & 65.44\% & -- \\
Shi et al.~\cite{shi_improved_2022} & -- & I/Q & -- & CNN, SE block & 113k & C & -- & 98.70\% \\
Harper et al.~\cite{harper_learnable_2024} & -- & -- & -- & CNN, SE block & 200k & C & 63.15\% & -- \\
Harper et al.~\cite{harper_automatic_2023} & -- & I/Q & -- & CNN, SE block & 203k & C & 63.70\% & 98.90\% \\
Huynh-The et al.~\cite{huynh-the_mcnet_2020} & MCNet & I/Q & -- & CNN & 220k & C & -- & 93\% (20 dB) \\
Nisar et al.~\cite{nisar_lightweight_2023} & -- & I/Q & -- & CNN, SE block & 253k & A & -- & 81\% (18 dB) \\
Wang et al.~\cite{wang_multitask_2024} & MCLNet & -- & -- & CNN & 83k & A & 61.8\% & -- \\
Xue et al.~\cite{xue_micnet_nodate} & MICNet & I/Q & -- & CNN & 186k & C & -- & 97.5\% (28 dB) \\
Wang et al.~\cite{wang_automatic_2023} & CVResNet & Random mix/rotate & -- & CNN & 308k & A & -- & 96.4\% (2 dB) \\
Xiao et al.~\cite{xiao_complex-valued_2023} & CDSCNN & I/Q & -- & CNN & 330k & A & 62.47\% & -- \\
 &  &  &  &  & 330k & B & 63.15\% & -- \\
Singh et al.~\cite{singh_automatic_2025} & -- & I/Q & -- & CNN & 336k & B & -- & 93.3\% (12 dB) \\
Duan et al.~\cite{duan_multi-modal_2023} & M-LSCANet & Constellation diagram (low SNR) & -- & -- & 46k ($<-4$ dB), 544k ($\ge -4$ dB) & B & 65.56\% & 96.4\% \\
Wei et al.~\cite{wei_adaptive_2023} & SCSNN & I/Q & $\checkmark$ & CNN & 580k & A & 63.1\% & -- \\
  &  &  &  &  & 580k & B & 64.9\% & -- \\
  &  &  &  &  & 580k & C & 65.0\% & -- \\
Cao et al.~\cite{cao_modulation_2023} & MobileNet & I/Q & -- & CNN & 628k & NS & -- & 82.2\% (30 dB) \\
Tekbiyik et al.~\cite{tekbiyik_robust_2020} & -- & I/Q & -- & CNN & 659k & A & -- & 90.7\% \\
An and Lee~\cite{an_robust_2023} & TADCNN & I/Q & $\checkmark$ & CNN & 671k & A & 66.64\% & 91\% \\
Guo et al.~\cite{guo_robust_2023} & KD-GSENet & KD-tree, enhanced constellation & -- & CNN,SE block & 10M & NS & -- & -- \\
Le et al.~\cite{le_performance_2022} & -- & FFT,STFT & -- & CNN & 12M & NS & 62.21\% & 65.18\% \\
Triaridis et al.~\cite{triaridis_mm-net_2024} & MM-Net & Constellation,diagram HOC & -- & -- & -- & NS & -- & 98.4\% (0--15 dB) \\
Wu et al.~\cite{wu_automatic_2020} & -- & I/Q & -- & -- & 50M & B & -- & 93.7\% (14 dB) \\

\midrule
\multicolumn{9}{c}{\textbf{RNN Models}}\\
\midrule
Ke and Vikalo~\cite{ke_real-time_2022} & DAE & A/P & -- & LSTM & 14k & C & 58.74\% & 97.91\% \\
Ansari et al.~\cite{ansari_novel_nodate} & -- & I/Q & -- & LSTM & 840k & NS & -- & 99.87\% (-5 dB) \\

\midrule
\multicolumn{9}{c}{\textbf{Transformer Models}}\\
\midrule
Li et al.~\cite{li_lightweight_2023} & CV-TRN & I/Q, random phase offset & -- & Complex-value transformer & 44k--254k & A & 63.70\%--64.43\% & 93.51\%--94.89\% \\
  &  &  &  & Complex-value transformer & 44k--254k & C & 63.60\%--65.09\% & 97.73\%--99.12\% \\
Su et al.~\cite{su_robust_2022} & SigFormer & I/Q & -- & Transformer & 44k & A & 63.71\% & 93.60\% \\
  &  &  &  &  & 44k & B & 65.77\% & 94.80\% \\
  &  &  &  &  & 158k & C & 63.96\% & 97.50\% \\
Zheng et al.~\cite{zheng_tmrn-glu_2022} & TMRN-GLU(small/large) & I/Q & -- & Transformer & 25k (Small), 106k (Large) & B & 61.7\% (Small), 65.7\% (Large) & 93.70\% \\

\midrule
\multicolumn{9}{c}{\textbf{SNN Models}}\\
\midrule
Lin et al.~\cite{lin_fast_2024} & -- & I/Q converted to coordinate rectangular system & -- & SNN & 7k & A & 36.39\% & -- \\
  &  &  &  &  & 7k & B & 39.74\% & -- \\
  &  &  &  &  & 7k & C & 53.79\% & -- \\
Guo et al.~\cite{guo_end--end_2024} & -- & SigmaDelta spiking encoding & -- & SNN & 84k--627k, 83k--542k & A & 56.69\% & -- \\
  &  &  &  &  & 84k--627k, 83k--542k & C & 64.29\% & -- \\

\midrule
\multicolumn{9}{c}{\textbf{Hybrid Models}}\\
\midrule
Suman and Qu~\cite{suman_lightweight_2025} & -- & I/Q, A/P & $\checkmark$ & CNN,LSTM,DP-DRSN & 27k & A & 61.20\% & 91.23\% \\
  &   &   &  &   & 27k & B & 63.78\% & 93.64\% \\
  &   &   &  &  & 27k & C & 62.13\% & 97.94\% \\
TianShu et al.~\cite{tianshu_iq_2022} & IQCLNet & I/Q & -- & CNN,LSTM & 29k & A & 59.73\% & -- \\
An et al.~\cite{an_efficient_2025} & TDRNN & I/Q & $\checkmark$ & CNN,GRU & 41k & A & 63.5\% (-8--18 dB) & -- \\
Gao et al.~\cite{gao_modulation_nodate} & -- & I/Q, A/P & -- & CNN,LSTM & 43k & A & 66.7\% (5 modulations) & -- \\
Ding et al.~\cite{ding_data_2023} & -- & I/Q, A/P & -- & CNN, BiGRU & 69k & C & -- & 34\% (-10 dB) \\
Zhang et al.~\cite{zhang_efficient_2021} & PET-CGDNN & I/Q & -- & CNN,GRU & 71k--75k & A & 60.44\% & -- \\
  &  &  &  &  & 71k--75k & B & 63.82\% & -- \\
  &  &   &   &   & 71k--75k & C & 63.00\% & -- \\
Xue et al.~\cite{xue_mlresnet_nodate} & MLResNet & I/Q & -- & CNN, LSTM & 115k & C & -- & 96.6\% (18 dB) \\
Riddhi et al.~\cite{riddhi_dual-stream_2024} & -- & I/Q, A/P & -- & CNN,GRU & 145k & B & 68.23\% (digital mod) & -- \\
Parmar et al.~\cite{parmar_dual-stream_2023} & -- & I/Q, A/P & -- & CNN,BiLSTM & 146k & B & 68.23\% & -- \\
Parmar et al.~\cite{parmar_deep_2024} & -- & I/Q & -- & CNN, LSTM & 155k & B & 63\% & -- \\
Chang et al.~\cite{chang_fast_2023} & FastMLDNN & I/Q, A/P & -- & CNN,Transformer & 159k & A & 63.24\% & -- \\
Luo et al.~\cite{luo_rlitnn_2024} & RLITNN & I/Q, A/P and FFT & -- & CNN,LSTM, Transformer & 181k & A & 63.84\% & -- \\
  &  &  &  &  & 181k & B & 65.32\% & -- \\
Sun and Wang~\cite{sun_novel_2023} & FGDNN & I/Q, A/P & -- & CNN,GRU & 253k & C & -- & 90\% (8 dB) \\
Yang et al.~\cite{yang_irlnet_2021} & IRLNet & I/Q, A/P & -- & CNN,LSTM & 318k & B & -- & $>$93\% (0--18 dB) \\
Qu et al.~\cite{qu_enhancing_2024} & TLDNN & A/P & -- & LSTM,Transformer & 243k & A & 63.4\% & 93.4\% \\
  &  &  &  &  & 243k & C & 63.4\% & 97.4\% \\
Cheng et al.~\cite{cheng_automatic_2024} & CVCNN-LSTM & I/Q & -- & Complex-value CNN,LSTM & 320k & A & 62.22\% & -- \\
Ying et al.~\cite{ying_convolutional_2023} & CTDNN & I/Q & -- & CNN,Transformer & 348k & A & 63.5\% & -- \\
Hou et al.~\cite{hou_signal_2023} & IQGMCL & I/Q & -- & CNN, BiLSTM & 406k & A & -- & 92.3\% ($>$0 dB) \\
  &  &  &  &  & 406k & B & -- & 93.3\% ($>$0 dB) \\
Liu et al.~\cite{liu_deep_2024} & DH-TR & I/Q & -- & CNN,Transformer,GRU & 460k & A & 62.81\% & -- \\
  &  &  &  &  & 460k & B & 65.41\% & -- \\
  &  &  &  &  & 460k & C & 63.48\% & -- \\
Ma et al.~\cite{ma_automatic_2023} & MasNet & HTCS graph & -- & CNN,Transformer & 550k & A & 63.31\% & 94.45\% \\
  &  &  &  &  & 550k & C & 52.65\% & 93.31\% \\
Gao et al.~\cite{gao_crosstlnet_nodate} & Mini-CrossTLNet & I/Q, A/P & -- & TCN,LSTM & 575k & A & 63.75\% & -- \\

\end{longtable}
\normalsize

\section{}\label{secB1}

\begin{figure}[H]
\centering

\begin{subfigure}[b]{0.48\textwidth}
\centering
\includegraphics[width=\linewidth]{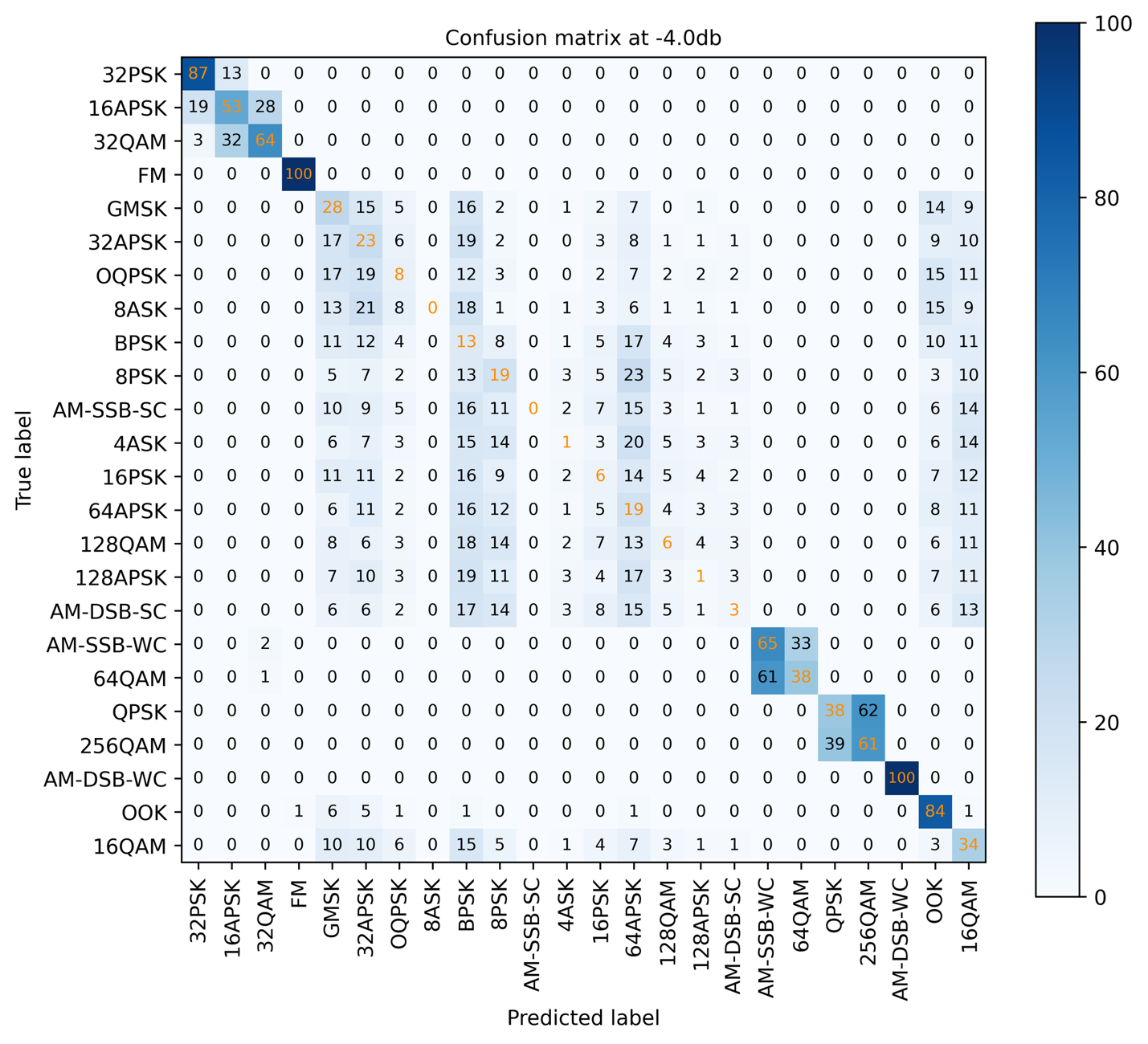}
\caption{SNR = -4 dB}
\label{fig:B1}
\end{subfigure}
\hfill
\begin{subfigure}[b]{0.48\textwidth}
\centering
\includegraphics[width=\linewidth]{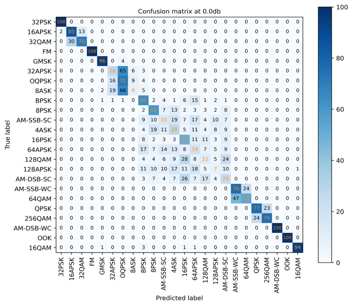}
\caption{SNR = 0 dB}
\label{fig:B2}
\end{subfigure}

\vspace{0.6em}

\begin{subfigure}[b]{0.48\textwidth}
\centering
\includegraphics[width=\linewidth]{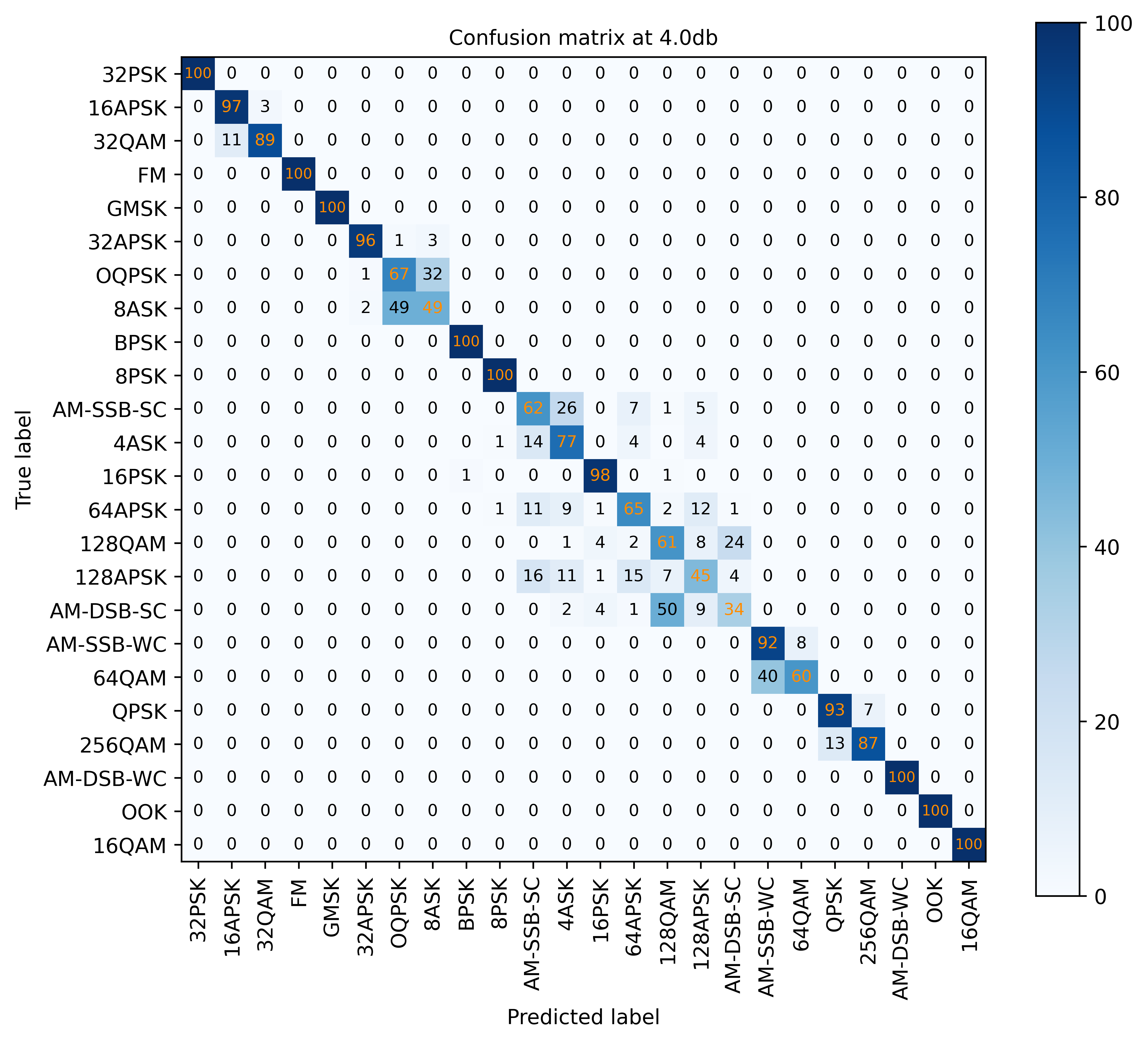}
\caption{SNR = 4 dB}
\label{fig:B3}
\end{subfigure}
\hfill
\begin{subfigure}[b]{0.48\textwidth}
\centering
\includegraphics[width=\linewidth]{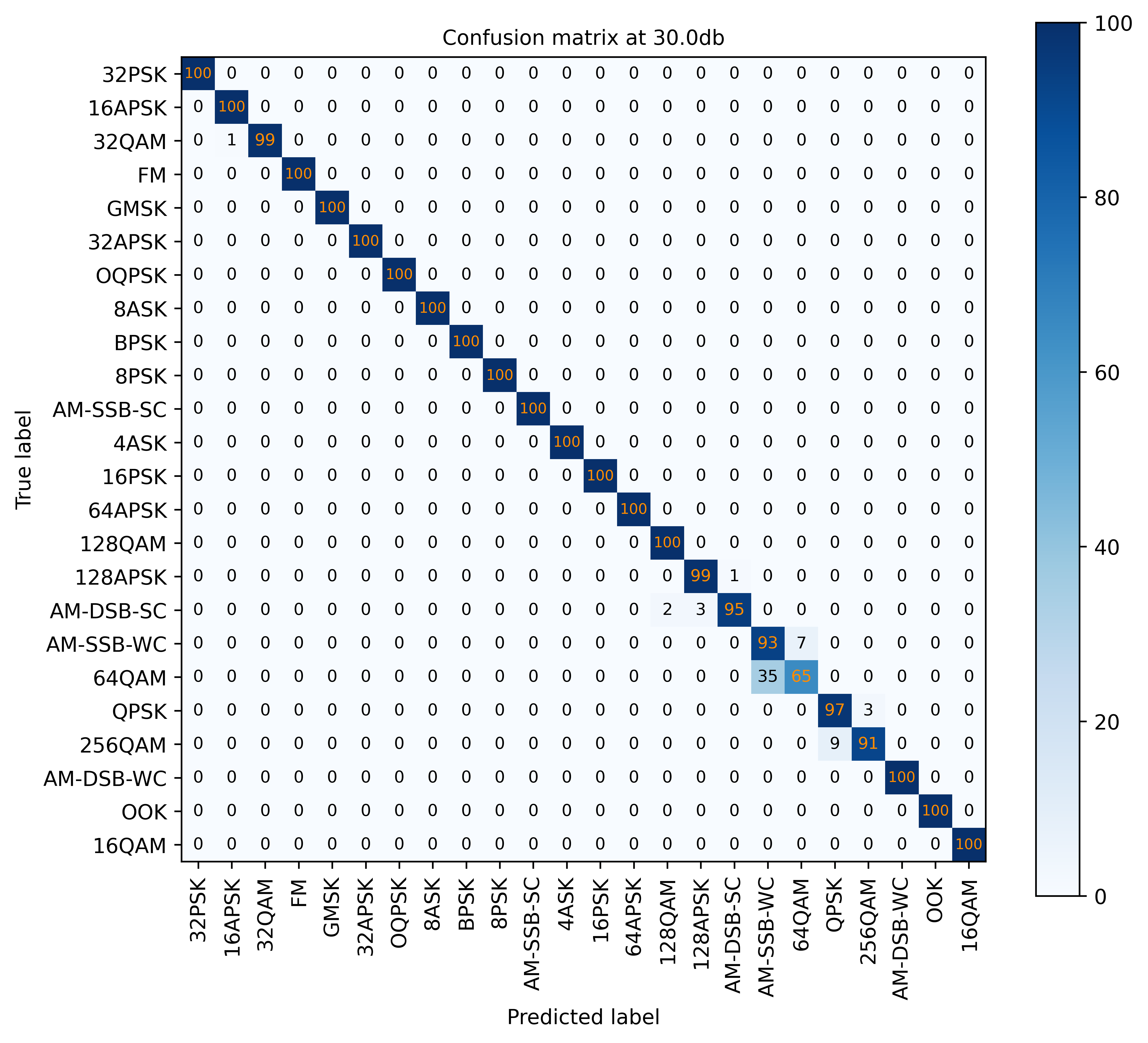}
\caption{SNR = 30 dB}
\label{fig:B4}
\end{subfigure}

\caption{Proposed model confusion matrices at -4dB, 0dB, +4dB, and +30dB SNR}
\label{fig:confmat_2x2}
\end{figure}

\clearpage
\section{}\label{secC1}

\captionsetup{type=table}
\captionof{table}{Proposed model vs benchmark models average modulation classification accuracy at each SNR level}
\label{tab:snr_accuracy}

\centering
\small
\begin{tabular}{ccccc}
\toprule
SNR Level (dB) & Proposed Model (Model 2) & PET-CGDNN & MCLDNN & DAE \\
\midrule
-20 & 4.219\% & 4.193\% & 3.867\% & 4.193\% \\
-18 & 4.538\% & 4.751\% & 4.059\% & 4.432\% \\
-16 & 4.697\% & 4.972\% & 5.419\% & 4.370\% \\
-14 & 6.397\% & 6.111\% & 5.578\% & 4.261\% \\
-12 & 8.964\% & 9.162\% & 7.705\% & 4.205\% \\
-10 & 13.747\% & 13.244\% & 11.799\% & 4.234\% \\
-8  & 20.336\% & 18.498\% & 16.805\% & 5.278\% \\
-6  & 28.414\% & 25.243\% & 25.238\% & 9.405\% \\
-4  & 34.946\% & 32.923\% & 34.998\% & 17.285\% \\
-2  & 46.408\% & 43.717\% & 44.517\% & 34.522\% \\
0   & 58.911\% & 54.903\% & 56.849\% & 47.617\% \\
2   & 71.500\% & 67.088\% & 68.244\% & 62.163\% \\
4   & 81.899\% & 75.509\% & 81.389\% & 72.481\% \\
6   & 91.688\% & 86.729\% & 90.671\% & 85.148\% \\
8   & 96.284\% & 93.011\% & 95.430\% & 91.720\% \\
10  & 96.973\% & 94.997\% & 96.882\% & 95.225\% \\
12  & 97.282\% & 95.936\% & 97.483\% & 96.952\% \\
14  & 97.859\% & 96.736\% & 97.542\% & 97.497\% \\
16  & 98.001\% & 96.948\% & 97.914\% & 97.685\% \\
18  & 97.269\% & 96.410\% & 97.832\% & 97.564\% \\
20  & 97.661\% & 96.530\% & 97.784\% & 97.545\% \\
22  & 97.664\% & 96.624\% & 97.901\% & 97.804\% \\
24  & 97.618\% & 96.741\% & 97.764\% & 97.395\% \\
26  & 97.917\% & 96.934\% & 97.998\% & 97.747\% \\
28  & 97.700\% & 96.756\% & 97.668\% & 97.906\% \\
30  & 97.472\% & 96.443\% & 97.791\% & 97.485\% \\
\bottomrule
\end{tabular}
\normalsize




\end{appendices}


\bibliography{references}

\clearpage
\section*{Figure Titles and Legends}

\noindent\textbf{Figure 1. Proposed model architecture}\\
This figure presents the overall design of the proposed automatic modulation classification model. The architecture combines two complementary feature extraction pathways. One branch uses a long short-term memory network to learn temporal dependencies across the signal sequence, while the second branch applies cross-channel self-attention to model instantaneous interactions between the in-phase and quadrature components at each time step. The outputs of both branches are fused and passed through a cascade of dual-path residual shrinkage denoising blocks with garrote thresholding. These denoising stages are designed to suppress noise while preserving modulation-discriminative information. The processed features are then aggregated and mapped to the final modulation class probabilities through a softmax classifier.

\vspace{0.5em}
\noindent\textbf{Figure 2. Cross-channel self-attention block}\\
This figure illustrates the proposed cross-channel self-attention mechanism for I/Q signal modeling. At each time step, the in-phase and quadrature components are treated as a two-token representation and projected into a latent feature space. Self-attention is then applied across the two tokens to capture their pairwise dependency structure. The block includes query, key, and value projections, scaled dot-product attention, multi-head aggregation, residual connection, layer normalization, and a feed-forward sublayer. By restricting attention to the two-channel representation instead of the full sequence, the design captures cross-channel coupling efficiently.

\vspace{0.5em}
\noindent\textbf{Figure 3. Denoising block A and B}\\
This figure shows the two denoising modules used in the proposed network. Denoising Block A performs adaptive thresholding on the incoming feature map using pooled feature statistics. Global average pooling and global max pooling are used to estimate complementary information about feature magnitude and distribution, and these statistics are combined through learnable coefficients to generate a data-driven threshold. Denoising Block B extends the same principle while also reducing spatial dimensions and increasing channel depth, enabling multiscale feature refinement. Together, the two blocks form the denoising backbone of the model and are intended to attenuate noise while retaining informative structures needed for modulation classification.

\vspace{0.5em}
\noindent\textbf{Figure 4. Garrote thresholding and its derivative}\\
This figure depicts the garrote thresholding function used in the denoising stages and the corresponding derivative used during optimization. The thresholding function suppresses coefficients with small magnitude while retaining and smoothly shrinking larger coefficients. Compared with more aggressive thresholding methods, garrote thresholding is intended to reduce over-smoothing and better preserve important modulation-related features. The derivative plot illustrates the smooth transition behavior above the threshold, which helps stabilize gradient-based learning during backpropagation. 

\vspace{0.5em}
\noindent\textbf{Figure 5. Proposed model architecture with tensor shapes}\\
This figure provides a detailed view of the proposed model architecture with tensor dimensions included at each major stage. The input signal is represented as a sequence of in-phase and quadrature samples. One branch extracts temporal information using a long short-term memory layer, while the second branch applies cross-channel self-attention to the two-channel representation. The attention output is projected, padded, and fused with the temporal branch output. The fused tensor then passes through denoising blocks, normalization, activation, global pooling, and final classification layers. The tensor annotations clarify how feature sizes evolve throughout the network and how the two branches are aligned before fusion.

\vspace{0.5em}
\noindent\textbf{Figure 6. Normalized modulation class distribution for train, validation, and test samples}\\
This figure shows the normalized distribution of modulation classes across the training, validation, and test subsets used in the study. The plot demonstrates that stratified sampling was applied so that each subset maintains a balanced representation of all modulation categories. This balanced partitioning helps reduce sampling bias and supports fair model training and evaluation. By preserving comparable class proportions across the three subsets, the experimental design ensures that performance differences are less likely to arise from uneven class exposure and more likely to reflect the true behavior of the proposed model.

\vspace{0.5em}
\noindent\textbf{Figure 7. Normalized SNR level distribution for train, validation, and test samples}\\
This figure shows the normalized distribution of signal-to-noise ratios across the training, validation, and test subsets. The plot confirms that stratified sampling was also maintained across SNR conditions, ensuring that low, moderate, and high-SNR samples are proportionally represented in each subset. This distribution is important because the study’s central objective is to improve robustness under noisy conditions. Maintaining similar SNR profiles across all subsets supports reliable comparison between training and evaluation results and helps prevent bias toward any specific noise regime.

\vspace{0.5em}
\noindent\textbf{Figure 8. Proposed model training and validation loss}\\
This figure shows the loss curves of the proposed model during training and validation. The trajectories provide a visual summary of optimization behavior across epochs and help assess convergence, generalization, and possible overfitting. A steady decline in training loss, alongside a stable validation loss, indicates that the model is learning discriminative representations without severe divergence between training and validation performance. The figure supports the claim that the selected optimization settings and stopping criteria yielded a stable training process for the proposed architecture.

\vspace{0.5em}
\noindent\textbf{Figure 9. Proposed model training and validation accuracy}\\
This figure shows the training and validation accuracy curves for the proposed model across training epochs. The curves are used to evaluate the consistency of learning progress and the degree of agreement between fitting on the training data and performance on held-out validation data. Similar trends between the two curves suggest stable learning and limited overfitting. The figure complements the loss analysis by showing how predictive performance evolves throughout training and by supporting the selection of the final trained model.

\vspace{0.5em}
\noindent\textbf{Figure 10. Proposed model modulation classification accuracy at each SNR level}\\
This figure reports the modulation classification accuracy of the proposed model as a function of signal-to-noise ratio. It illustrates how performance changes from highly noisy conditions to cleaner signal conditions. The plot is particularly important because the study focuses on improving robustness at low and moderate SNR values, where many automatic modulation classification models degrade substantially. The curve shows that accuracy improves progressively with increasing SNR while maintaining useful performance in more difficult low-SNR regions, demonstrating the effectiveness of the feature-preserving denoising design.

\vspace{0.5em}
\noindent\textbf{Figure 11. Proposed model vs benchmark models modulation classification accuracy at each SNR level}\\
This figure compares the proposed model with benchmark models across signal-to-noise ratio levels. The comparison highlights the relative classification accuracy achieved by the proposed architecture and shows where the largest gains occur. Particular attention is given to the low to moderate SNR region, where the proposed model is intended to provide greater robustness through cross-channel attention and adaptive denoising. The figure visually supports the reported improvement over PET-CGDNN, MCLDNN, and DAE under identical evaluation conditions and illustrates the practical advantage of the proposed design in challenging noise environments.

\vspace{0.5em}
\noindent\textbf{Figure 12. Ablation study: Average modulation classification accuracy vs. FLOP; color by inference time/sample (ms)}\\
This figure summarizes the trade-off between average classification accuracy and computational cost for the ablation models. Each point represents one model configuration, with computational complexity measured in floating-point operations and inference latency encoded by color. The plot is intended to show how architectural changes influence both predictive performance and efficiency. It highlights the region where accuracy improves as complexity increases and the point at which further scaling yields only marginal gains. This visualization supports the conclusion that the best model is not the largest one, but rather the configuration that balances denoising depth, attention depth, and temporal modeling most efficiently.

\vspace{0.5em}
\noindent\textbf{Figure 13. Ablation study: average modulation classification accuracy per SNR level}\\
This figure compares the modulation classification accuracy of the ablation models across SNR levels. The curves illustrate how different architectural choices affect robustness under varying noise conditions. By displaying performance as a function of SNR, the figure makes it possible to identify which design changes contribute most strongly to low-SNR resilience and which changes mainly affect high-SNR behavior. The comparison supports the finding that denoising depth has the largest influence on low-SNR performance, while moderate cross-channel self-attention depth provides secondary but consistent gains.

\noindent\textbf{Figure B1. Proposed model confusion matrices at -4dB, 0dB, +4dB, and +30dB SNR}\\
This figure presents the confusion matrices of the proposed automatic modulation classification model at four representative signal-to-noise ratio levels: $-4$ dB, $0$ dB, $+4$ dB, and $+30$ dB. Each matrix shows the relationship between true modulation labels and predicted labels, thereby illustrating how class-level decision behavior changes as signal quality improves. At lower SNR values, the matrices show broader dispersion of predictions and more frequent confusion among modulation classes with similar temporal or spectral characteristics. The sequence of matrices provides a class-wise view of robustness that complements the aggregate accuracy curves reported in the main results. Together, these panels demonstrate that the proposed model gains discrimination power as noise decreases while still retaining meaningful recognition capability in challenging low-SNR conditions.

\end{document}